\documentclass[sn-mathphys-num]{sn-jnl}

\usepackage{graphicx}%
\usepackage{mathrsfs}%
\usepackage[title]{appendix}%
\usepackage{textcomp}%
\usepackage{manyfoot}%
\usepackage{algorithmicx}%
\usepackage{algpseudocode}%
\usepackage{listings}%

\usepackage{amsmath}
\usepackage{amssymb}
\usepackage{mathtools}
\usepackage{amsthm}

\usepackage[utf8]{inputenc} 
\usepackage[T1]{fontenc}    
\usepackage{hyperref}       
\usepackage{url}            
\usepackage{booktabs}       
\usepackage{amsfonts}       
\usepackage{nicefrac}       
\usepackage{microtype}      
\usepackage[dvipsnames]{xcolor}
\usepackage{natbib}
\usepackage[ruled]{algorithm2e}
\usepackage{float}
\usepackage{makecell}

\usepackage{bm}
\usepackage{multirow}
\usepackage{colortbl}
\usepackage{enumitem}
\usepackage{framed}
\usepackage{tcolorbox}
\usepackage{setspace}
\usepackage{array}
\usepackage{wrapfig}
\usepackage{tabularx}
\usepackage{subcaption}
\usepackage{utfsym}

\usepackage[capitalize,noabbrev]{cleveref}

\renewcommand{\paragraph}[1]{\noindent\textbf{#1}\quad}

\theoremstyle{thmstyleone}%
\theoremstyle{thmstyletwo}%

\theoremstyle{thmstylethree}%

\begin{document}

\title[]{\centering Humanoid-inspired Causal Representation Learning for Domain Generalization}


\author[1]{\fnm{Ze} \sur{Tao}}\email{taoze.tz@csu.edu.cn}
\equalcont{These authors contributed equally to this work.}

\author*[1]{\fnm{Jian} \sur{Zhang}}\email{jianzhang@csu.edu.cn}
\equalcont{These authors contributed equally to this work.}

\author[1]{\fnm{Haowei} \sur{Li}}\email{234711031@csu.edu.cn}

\author[1]{\fnm{Xianshuai} \sur{Li}}\email{254701004@csu.edu.cn}

\author[1]{\fnm{Yifei} \sur{Peng}}\email{hallzerk@csu.edu.cn}

\author*[1]{\fnm{Xiyao} \sur{Liu}}\email{lxyzoewx@csu.edu.cn}

\author[1]{\fnm{Senzhang} \sur{Wang}}\email{szwang@csu.edu.cn}

\author[2]{\fnm{Chao} \sur{Liu}}\email{chao-liu23@mails.tsinghua.edu.cn}

\author[3]{\fnm{Sheng} \sur{Ren}}\email{rensheng@huas.edu.cn}

\author[4]{\fnm{Shichao} \sur{Zhang}}\email{zhangsc@gxnu.edu.cn}


%
%
%
%
%
%
%

\affil[1]{\orgdiv{School of Computer Science}, \orgname{Central South University}, \orgaddress{\street{No.932 South Lushan Road}, \city{Changsha}, \postcode{410083}, \state{Hunan}, \country{China}}}

\affil[2]{\orgdiv{School of Computer Science}, \orgname{Tsinghua University}, \orgaddress{\street{No.30 Shuangqing Road}, \city{Beijing}, \postcode{100084}, \country{China}}}

\affil[3]{\orgdiv{School of Computer and Electrical Engineering}, \orgname{Hunan University of Arts and Sciences}, \orgaddress{\street{No. 3150, Dongting Road}, \city{Changde}, \postcode{415000}, \state{Hunan}, \country{China}}}

\affil[4]{\orgdiv{Guangxi Key Lab of Multi-Source Information Mining \& Security}, \orgname{Guangxi Normal University}, \orgaddress{\street{No. 15, Yucai Road, Qixing District}, \city{Guilin}, \postcode{541004}, \state{Guangxi}, \country{China}}}

\abstract{This paper proposes the Humanoid-inspired Structural Causal Model (HSCM), a novel causal framework inspired by human intelligence, designed to overcome the limitations of conventional domain generalization models. Unlike approaches that rely on statistics to capture data-label dependencies and learn distortion-invariant representations, HSCM replicates the hierarchical processing and multi-level learning of human vision systems, focusing on modeling fine-grained causal mechanisms. By disentangling and reweighting key image attributes such as color, texture, and shape, HSCM enhances generalization across diverse domains, ensuring robust performance and interpretability. Leveraging the flexibility and adaptability of human intelligence, our approach enables more effective transfer and learning in dynamic, complex environments. Through both theoretical and empirical evaluations, we demonstrate that HSCM outperforms existing domain generalization models, providing a more principled method for capturing causal relationships and improving model robustness. The code is available at \url{https://github.com/lambett/HSCM}.}

\keywords{Deep Learning; Causal Learning; Domain Generalization; Feature Representation}



\maketitle

\backmatter

\section{Introduction}
\label{sec:intro}
In recent years, the out-of-distribution (OOD) problem \cite{yang2024generalized} has highlighted the limitations of traditional deep learning models, especially under the i.i.d. assumption, when faced with new or unseen data distributions. Unlike neural networks, human intelligences can easily adapt to new tasks by leveraging prior knowledge and experience \cite{martinez2024decomposing}. The human visual system integrates features like shape, motion, and texture to form causal understanding. In contrast, neural networks often overlook the causes of domain shifts, limiting generalization. Domain generalization (DG) \cite{wang2022generalizing} aims not only to capture diverse features but to interpret them causally, enabling models to generalize beyond surface-level sensory input.


Recently, many remarkable methods have emerged in the DG field, focusing on improving the transferability of learned representations by strengthening statistical dependencies between the source and target domains. These methods typically introduce two regularization terms: one \cite{carlucci2019domain, volpi2018generalizing, zhou2020deep, zhou2020learning} to enhance the statistical dependence between representations and labels in the target domain, and another \cite{balaji2018metareg, li2019episodic, chattopadhyay2020learning, piratla2020efficient, dou2019domain, wang2020learning} to improve the relationship between the representations in the source and target domains. However, these statistical techniques often fail to distinguish properly between style and content in both domains, leading to two main issues: (i) learning spurious correlations between style and labels, and (ii) neglecting the true independence between causal content and labels. As a result, while these methods do improve transferability, they still face critical challenges, particularly in terms of effectively separating content from style and enhancing causal representation.

\begin{figure}[h]
    \centering
    \includegraphics[width=0.9\columnwidth]{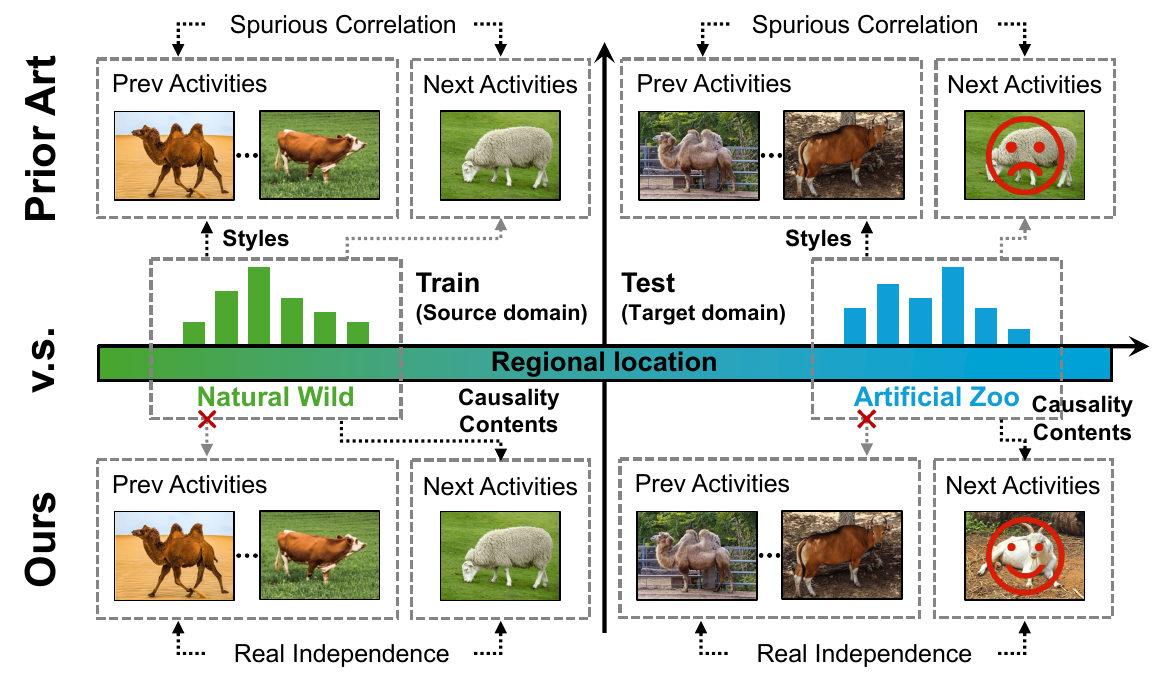}
    \setlength{\abovecaptionskip}{0.0cm}
	\setlength{\belowcaptionskip}{0.0cm}
    \vspace{-0.00em}
    \caption
	{
A toy example in a classification model distinguishing between natural wild and artificial zoos. Prior research often spuriously correlates non-causal factors, which can lead to misleading activities. Our approach seeks to address this issue by counteracting the effects of style-based confounding factors, such as desert and grassland’s color, which are often co-occurring with ground-truth like camels, cows, and sheep in the wild.
    }
    \label{fig1}
    \vspace{-0.00em}
\end{figure}


Numerous causal DG methods \cite{chen2023meta, lv2022causality} construct causal graphs to model variable relationships and identify true independencies. Although causal inference alleviates some challenges, most methods rely on uniformly encoded low-order visual cues. Such coarse structures often fail under large visual discrepancies between source and target domains. As the upper part of Fig. \ref{fig1}, without further disentangling style and content, these methods may lead to misinterpretations. In such cases, they might primarily focus on optimizing objective results, such as the strong association between styles (e.g., desert and grassland’s color) in the source domain and labels (e.g., camels, cows, and sheep) in the target domain, rather than the essential content features, creating a gap with human subjectivity. This results in overfitting to domain-specific features, such as background or illumination, rather than focusing on the relevant content features, such as the shape and edge. The ideal methods should reflect the human ability to extract visual information across different environments, which requires the model to fine-grained learn styles (e.g., color and texture) and contents (e.g., shape) in the source domain (e.g., the wildlife park) and iteratively aggregate and develop complementary information to facilitate high-order causal factors in the target domain (e.g., the zoo). Accurately estimating causal factor importance requires modeling their influence distributions on decisions. However, such information is often inaccessible due to data limitations, forcing models to rely solely on observed data. This highlights the need for causal theories that adapt to unobserved contexts while aligning with human-like subjective understanding.


In this paper, we integrate the mechanisms of human visual perception (HVP) into causal-based DG and propose a novel framework. As illustrated in the lower part of Fig. \ref{fig1}, our framework explicitly separates the influences of various visual attributes, preventing the DG model from being misled by false correlations and independencies. Specifically, we decouple unobservable and naturally mixed contents and styles, aligning with the HVP. The decoupling allows the model to better establish a hierarchical processing and multi-level learning structure in the Structural Causal Model (SCM). We then employ data transformations that simulate the impact of contextual factors on the data generation process, using causal intervention for SCM to optimize transferability. Additionally, to handle confounding effects across environments, we design an adaptive strategy that adjusts the number and weights of transformations based on their importance, dynamically selecting effective representations and evolving the architecture to fit varying contexts. Our contributions can be summarized as follow:

\begin{itemize}
    \item The novel HSCM framework theoretically formalizes the fine-grained relationship between style, content, and labels in causal-based DG. It re-identifies the causal dependencies from low-order to high-order components, leveraging hierarchical human visual perception mechanisms to model dependencies between data, color, texture, shape, and labels.
    \item The HSCM framework simulates data generation, information processing, and decision-making processes in human cerebral cortexes. By integrating causal inference with high-order visual attributes of human vision, HSCM stands as a pioneering approach, investigating synergistic correlations between causal factors.
    \item Experiments on multiple DG benchmarks show that the proposed framework consistently outperforms state-of-the-art methods. Moreover, its interpretability is demonstrated in real-world scenario tasks.
\end{itemize}

\section{Related Work}
\label{section:B}

\subsection{Humanoid-inspired learning}
Humanoid-inspired learning improves robustness and interpretability in computer vision by replicating the hierarchical processing and multimodal learning of the human vision. It focuses on rapidly learning and transferring knowledge, adapting to complex and ever-changing external environments. For instance, some methods \cite{ge2022contributions, wang2024fusion} explicitly disentangles visual features (color, texture, shape, depth, motion), providing a more systematic understanding of the contributions of each feature. These features remain stable and invariant under environmental transformations, making them particularly effective in tackling challenges such as domain shifts and adversarial robustness. Moreover, a key aspect of humanoid-inspired learning is its ability to incorporate insights from cognitive processes, such as attention mechanisms. These mechanisms primarily consist of two processes: bottom-up, unconscious guidance \cite{shi2022visual, yang2025dashboard} and top-down, task-driven guidance \cite{chen2024frequency, wang2025mmae, denison2024visual}. The bottom-up mechanism automatically directs attention to prominent features in an image, such as color, brightness, and contrast, while the top-down mechanism actively focuses attention on relevant areas based on the task at hand. Different regions of an image compete for visual attention, with only those exhibiting strong visual stimuli—known as salient regions—able to capture attention. This mechanism enables the model to mimic the way humans selectively focus on key visual information, improving both the efficiency and accuracy of classification.
%

\subsection{Domain Generalization}
DG aims to improve model performance in unseen domains by learning representations that are robust to distribution shifts. Because DG does not assume access to target domain data during training, making it more challenging yet crucial for real-world applications. Early DG methods \cite{zhou2022domain, cubuk2020randaugment, cubuk2019autoaugment} reduced domain-specific biases by augmenting the data space with techniques such as geometric transformations, color adjustments, noise injection, and image fusion, thereby improving the model’s generalization ability. Later methods focused on feature alignment strategies, with invariant feature learning aiming to find a mapping function that minimizes the distance between domains in the feature space \cite{xu2019d}, often achieved through adversarial training \cite{fan2021adversarially, zhao2020maximum, volpi2018generalizing} or kernel-based domain-invariant representations \cite{koltchinskii2011oracle}. Feature disentanglement, another key area in DG research, suggests that the feature space can be decomposed into domain-invariant and domain-specific components. By decoupling these features, the impact of domain-specific elements on domain generalization can be minimized. Techniques like self-supervised learning \cite{carlucci2019domain, motiian2017unified} and feature augmentation \cite{wang2021learning, chattopadhyay2020learning, volpi2019addressing} further enrich feature representations, enhancing resilience to domain-specific shifts. Additionally, bias mitigation plays a crucial role in DG, as models often rely on domain-dependent shortcuts that hinder generalization. Strategies like meta-learning \cite{jain2024improving} simulate domain shifts during training, forcing the model to adapt to diverse conditions. Re-weighting \cite{dou2019domain} methods also help reduce over-reliance on spurious correlations by emphasizing invariant and discriminative features. These approaches align with DG’s overarching goal of achieving robust performance without compromising adaptability.
%

\subsection{Causal Inference for Domain Generalization}

Causal inference has become a key approach for addressing challenges in DG by mitigating spurious correlations and domain-specific biases. A significant direction in causal DG is causal representation learning, which disentangles stable causal features from non-causal ones. For example, Structural Causal Models (SCMs) can isolate factors like object shape while reducing the influence of confounders such as lighting or background texture, providing a robust foundation for generalization. By modeling the data generation process through SCMs, methods \cite{chen2023meta, lv2022causality, mahajan2021domain} can identify causal relationships that generalize well across domains. Counterfactual reasoning \cite{lin2024towards, jiao2025domain} plays a crucial role in causal DG by simulating domain shifts and identifying invariant features, with counterfactual data augmentation generating hypothetical examples by altering non-causal factors, helping models prioritize causal features during training. Another key aspect is incorporating causal constraints into model training. Techniques like causal regularization, which enforces independence between features and domain-specific attributes, ensure that learned representations are invariant to external confounders. Optimization frameworks based on causal principles, such as progressive learning \cite{li2021progressive} and risk minimization \cite{mo2025domain}, further improve robustness. However, the methods mentioned above often struggle with generalization under unseen conditions due to their reliance on statistical associations in training data. In contrast, our method, inspired by human visual processing, emphasizes an iterative and progressively refined approach, which is valuable for causal inference in complex data analysis. By adjusting and optimizing the model layer by layer, our method reduces spurious correlations, clarifies the causal chain, and improves the ability to capture true causal relationships, supporting more accurate reasoning and decision-making.

\section{Problem and Task Instantiation}

Our goal is to causally infer the SCM for DG while aligning it with the human visual system (HVS). We achieve this using observational data to model invariant representations bridging source and target domains. In a classification task, let $X \in \mathbb{R}^{n \times d}$ denote $n$ i.i.d. low-dimensional image samples with label $Y \in [M]$, influenced by unobserved confounders $Z$ (e.g., domain variables). Source and target data are drawn from respective domain distributions $e_{so}, e_{tr} \subset e$. For example, the source dataset $\mathcal{D}_{so} = \{(x_i^{so}, y_i^{so})\}_{i=1}^{N^{so}}$ is sampled from $p(X, Y \mid e_{so})$.

\begin{figure}[t]
    \centering
    \includegraphics[width=0.85\columnwidth]{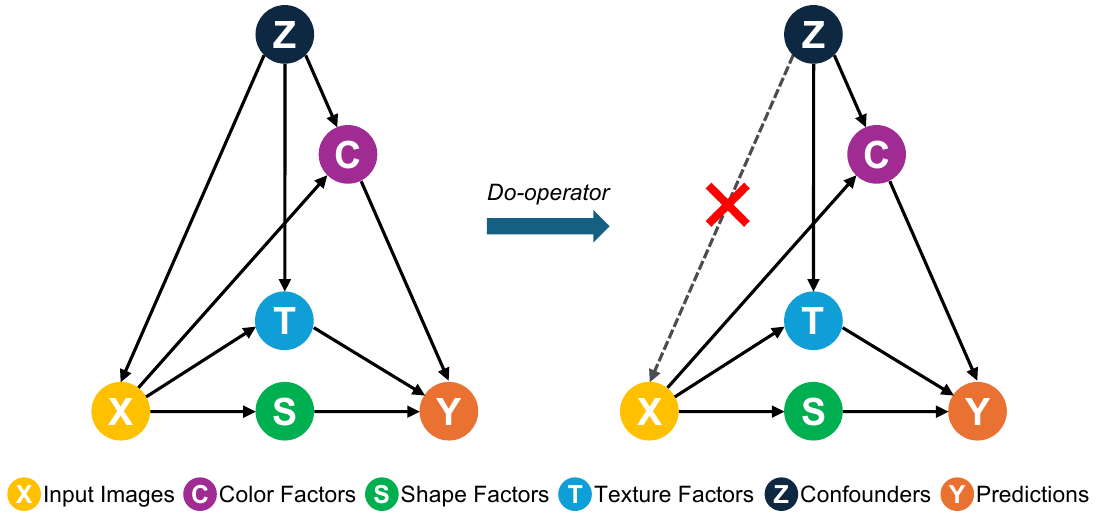}
    \setlength{\abovecaptionskip}{0.0cm}
	\setlength{\belowcaptionskip}{0.0cm}
    \vspace{-0.00em}
    \caption
	{
		 Structure causal graph of Humanoid-inspired Structural Causal Model (HSCM).
    }
    \label{fig2}
    \vspace{-0.00em}
\end{figure}

\subsection{Causal View at HSCM Task}

We begin by formulating a custom SCM to summarize the HSCM framework. As shown in Fig. \ref{fig2}, we assume the data generation process is represented by the SCM, which characterizes the observational data distribution \( p(X, Y) \). The discrepancies between the source and target domains, \( p_s \neq p_t \), are attributed to three observable causal variables: texture \( T \), color \( C \), and shape \( S \). This aligns with the observation that the HVS relies on these variables for classification. The DG causal graph involves six variables: input images \( X \), \( T \), \( C \), \( S \), confounder \( Z \), and predictions \( Y \), with causal relationships described as follows:

$Z \rightarrow X$. \( Z \) represents typically unobserved confounders that cannot be directly formulated. In classification tasks, different object categories within the dataset are often presented across various scenes, which may introduce contextual biases. For example, pet cats predominantly appear in indoor environments, while individuals are often assigned distinct characteristics (such as being robust or slender) based on subjective preferences. These contextual biases, represented by \( Z \), are inherently difficult to distinguish from the naturally entangled contents in both \(\mathcal{D}_{so}\) and \(\mathcal{D}_{tr}\). As a result, models may mistakenly assume false independence between contents and labels, complicating causal inference and negatively affecting model performance, i.e., \( Z \rightarrow X \).

$Z \rightarrow C \rightarrow Y$ and $Z \rightarrow T \rightarrow Y$. \( C \) and \( T \) are often considered essential distinguishing factors in visual tasks. However, in cross-domain tasks, these factors are highly susceptible to contextual variations, such as environment changes, represented by \( Z \rightarrow C \) and \( Z \rightarrow T \). In this causal pathway, the features learned from the source domain cannot be directly generalized to the target domain. As a result, when the model relies on these factors for prediction, the discrepancy between domains may lead to unreliable predictions, i.e., \( C \rightarrow Y \) and \( T \rightarrow Y \). For example, the color and texture differences between cartoons and sketches.

$X \rightarrow C \rightarrow Y$ and $X \rightarrow T \rightarrow Y$. Although \( C \) and \( T \) may introduce unreliable contextual information, the human visual system is highly sensitive to variations in these factors, such as the differences between red and green apples or the distinction between wooden and fabric sofas. In the causal graph, by establishing causal relationships \( X \to C \) and \( X \to T \), the model can independently extract color and texture representations from the input image \( X \), and subsequently concatenate the resulting feature vectors to support final classification. This causal pathway not only integrates multiple causal factors but also enhances the model's adaptability across different domains.

$X \rightarrow S \rightarrow Y$. \( S \) is critical for describing the boundaries and contours of object categories. Under domain shift, \( S \) exhibits relatively less variation compared to \( C \) and \( T \), making it a more reliable feature. This is because shape is an intrinsic property of an object, typically unaffected by external factors such as lighting or changes in angle. By leveraging \( S \), the model can mitigate interference from unreliable contextual semantics, i.e., \( Z \rightarrow S \), thus avoiding the learning of domain-specific bias features that do not contribute to the core task.

Understanding confounding effects is key to building reliable causal models. In this pathway, factor $Z$ reduces classification generalization by introducing dependencies between variables, so the model learns $P(Y \vert X_c, Z_c)$ instead of the desired $P(Y \vert X_c)$. Since $Z$ affects both $C$ and $T$ ($X \leftarrow Z \rightarrow C, T$), the model becomes sensitive to shifts in $Z$. When $Z$’s distribution changes, discrepancies such as $P(Y \vert X_c, Z_c^{so}) \neq P(Y \vert X_c, Z_c^{tr})$ arise, leading to biased inferences and degraded generalization.

\subsection{Causal Intervention via HSCM}
We denote the Directed Acyclic Graph (DAG) by \( G = (V(G), E(G)) \), where \( V(G) \in \{X, C, T, S, Z, Y\} \) represents the set of variables, and \( E(G) \) contains the causal relationships between variables in the SCM. To understand the specific significance of each variable in \( G \), we utilize the Markov property to simplify this complexity, based on the structural characteristics of the graph. Thus, the probability density function of the joint distribution can be defined as \( p(v) = \prod_{i}^d p\left(v_i \mid v_{pa(i)}\right) \), where \( pa(i) = \left\{ j \in V(G) : j \rightarrow i \in E(G) \right\} \) represents the set of parent nodes of \( i \) in \( G \), and \( d \) is the set composed of \( X_C \), \( X_T \), and \( X_S \). The term \( p(v_i \mid v_{pa(i)}) \) denotes the conditional probability density function of \( v_i \), given its parent nodes \( v_{pa(i)} \). Formally, we can express the SCM depicted in Fig. \ref{fig2}, along with its data generation process, as follows:
\begin{equation}
\begin{aligned}
X_c &= f_c(p(c), N_c) = f_c\left(\prod_{i=1}^{d} p\left(v_{c} \mid x_{pa(c)}\right), N_c\right), \\
X_t &= f_t(p(t), N_t), \quad X_s = f_s(p(s)),\\
\end{aligned}
\label{eqn3}
\end{equation}
where \( N \) represents the corresponding noise variables, which are assumed to be independent of each other and are typically ignored in the modeling process. Each \( f \) denotes the corresponding data generation function. Consequently, we can treat \( X \) as being generated from a combination of multiple balanced marginal distributions, which can be defined as:
\begin{equation}
\begin{aligned}
X &= f_x(p(x),N_x)\doteq\mathbf{f}_y(c, t, s) + \epsilon,
\end{aligned}
\end{equation}
here, with a slight abuse of notation, \( \epsilon \) represents the set of noise variables. Suppose \( p_\epsilon > 0 \), the above equation can be further expressed as:
\begin{equation}
\begin{aligned}
& p_{\mathbf{f}}(X \mid c, t, s, y) = p_\epsilon\left(X - \mathbf{f}_y(c, t, s)\right), \\
& p_{e_{so}}(X, Y, c, t, s) = p_{\mathbf{f}}(X \mid c, t, s, Y) p_{e_{so}}(c, t \mid Y) \\
& \quad \quad \times p(s \mid Y) p_{e_{so}}(Y) \mid p_{e_{so}}(c, t \mid Y), p_{e_{so}}(Y) > 0.
\end{aligned}
\end{equation}

Based on the above formulation, the causal intervention via HSCM primary goals are as follows:
\\\

\noindent \textbf{Goal 1.} \emph{Training a classification model \( f: X \rightarrow Y \) that maintains consistent performance across both the target domain \( D_{tr} \) and the source domain \( D_{so} \). This ensures the model generalizes effectively to unseen target domain data while leveraging the source domain data for training.}
\\

\noindent \textbf{Theorem 1.} \emph{For a given classifier \( f(X) \) with its cross-entropy \( L(p(Y \mid X)) = -\mathbb{E}_{p(X, Y)} \log p(Y \mid X) \), there always exists \( e_{tr} \), and it is possible to construct a new environment \( e_{H} \) such that the following equation holds consistently (see appendix materials for the proof):}
\begin{equation}
\exists e_{tr} \text { s.t. } L_{e_{tr}}\left(p_{e_{so}}(Y \mid X)\right)-L_{e_{tr}}\left(p_{e_{H}}(Y)\right)>0.
\end{equation}

\noindent \textbf{Goal 2.} \emph{Learning a SCM $\left(P_X, G\right)$, where \( P_X \) represents the data distribution and \( G \) is the causal directed acyclic graph (DAG). This SCM is designed to reveal the data transformation process from the source domain \( D_{so} \) to the target domain \( D_{tr} \), helping to capture the invariant causal structure between the domains.}
\\

\noindent \textbf{Theorem 2.} \emph{Consider a structural equation model (SEM) in Eq. \ref{eqn3} that is linear. Let \( X \in \mathbb{R}^{n \times d} \) denote the data matrix, \( B \in \mathbb{R}^{d \times d} \) be the weighted adjacency matrix with \( B_{ii} = 0 \) (i.e., no self-loops), and \( N \sim \mathcal{N}(0, \Sigma) \) be the noise matrix with variances \( \sigma_1^2, \sigma_2^2, \dots, \sigma_d^2 \) along its diagonal, \( \Upsilon(G) \) be the acyclicity constraint term with a coefficient \( \lambda \). As the data size increases, i.e., \( n \rightarrow \infty \), the Evidence Lower Bound (ELBO) transformed causal discovery optimization problem \( S(X, Y, G) \) under negative log-likelihood loss consistently holds, and the reweighted score function \( S_w(X,Y,G) \) is satisfied (see appendix materials for the proof):}
\begin{equation}
\begin{aligned}
& \underset{G}{\arg \min }\left\{S_w(X,Y,G) + \lambda \Upsilon(G)\right\} - \\
& \underset{G}{\arg \min }\left\{S(X,Y,G) + \lambda \Upsilon(G)\right\} \xrightarrow{\text{a.s.}} 0.
\end{aligned}
\end{equation}

\begin{figure*}[t]
    \centering
    \includegraphics[width=1.0\columnwidth]{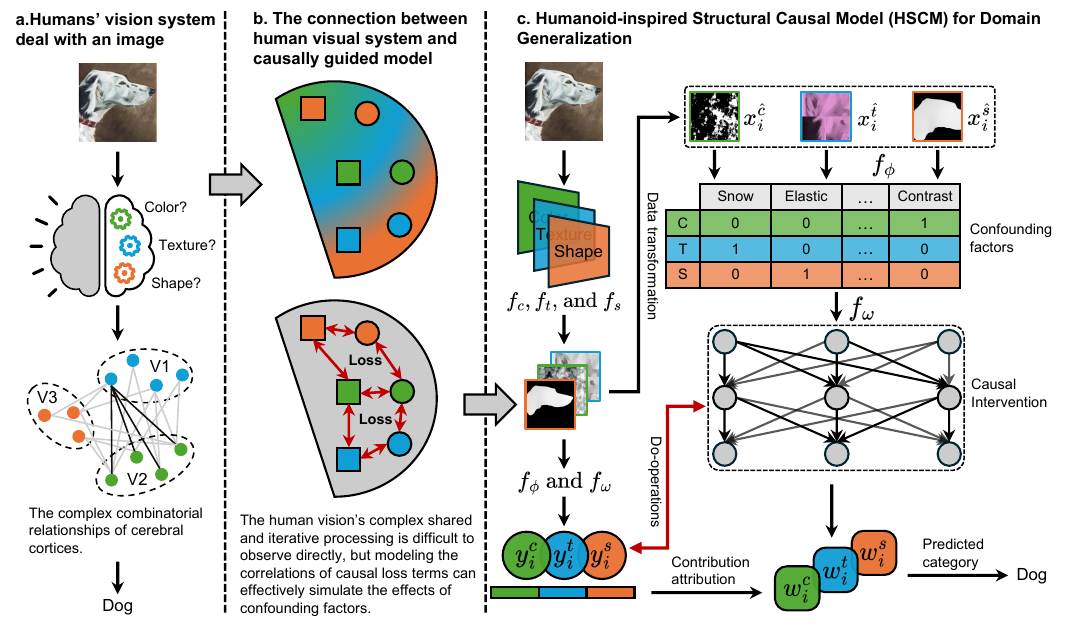}
    \setlength{\abovecaptionskip}{0.0cm}
	\setlength{\belowcaptionskip}{0.0cm}
    \vspace{+0.50em}
    \caption
	{
The proposed HSCM is outlined as: (a) Inspired by the iterative and shared processing of visual information in the cortical regions, we designed the HSCM framework as shown in (b)-(c). This framework sequentially extracts and analyzes visual attributes such as color, texture, and shape to simulate the human brain's visual processing mechanism.
    }
    \label{fig3}
    \vspace{-1.00em}
\end{figure*}
%


The proof above establishes the minimax optimality of the classifier trained on the HSCM and the identifiability of the DAG. This ensures robust model performance while maintaining DAG structure stability through an adaptive reweighting mechanism and acyclicity penalty.

\subsection{HSCM Task Instantiations}
After proving the invariance of HSCM, we disentangle the color \( Z_c \), texture \( Z_t \), and shape \( Z_s \) factors from the image \( X \) to obtain the corresponding features \( X_c \), \( X_t \), and \( X_s \). By reweighting their data transformations, we guide the causal representation learning process, prioritizing the most informative factors. This mechanism enhances the model’s ability to reduce the impact of spurious correlations. The process is formulated as:
\begin{equation}
X_c = f_c(X, Z_c), X_t = f_t(X, Z_t), X_s = f_s(X),
\end{equation}
here, \( Z_c \), \( Z_t \), and \( Z_s \) represent mutually independent noise terms corresponding to color, texture, and shape, respectively. Shape \( S \) is a stable factor correlated with the label \( Y \), while color \( C \) and texture \( T \) are environment-dependent noise factors. The components \( f_c(\cdot) \), \( f_t(\cdot) \), and \( f_s(\cdot) \) in HSCM serve as feature extractors for each factor, with their instantiation defined as follows:

\textbf{Color Feature Extractor.} We apply mutual perturbation to generate \( X_c \) from \( X \). First, we use the Fast Fourier Transform (FFT) to decompose the image into two components: the phase component and the magnitude component. Then, we randomly adjust the phase component to disrupt shape information while keeping color information unchanged. Next, we apply the Inverse Fast Fourier Transform (IFFT) to reconstruct the image, producing \( X_c \), which retains the original pixel color distribution but alters shape information. This approach effectively disentangles color and shape features, allowing for independent color feature analysis. Formally, the Fourier transform of a sample \( X \) can be represented as:
\begin{equation}
\mathcal{F}\left(x\right)=\mathcal{A}\left(x\right) \times e^{\mathcal{P}\left(x\right)},
\end{equation}
where \( \mathcal{A}(X) \) and \( \mathcal{P}(X) \) denote the amplitude and phase components of the image, respectively. Replacing the original phase \( \mathcal{P}(X) \) with \( \Phi \) and applying the IFFT to the modified frequency spectrum \( \mathcal{F}_{\text{per}}(X) \) reconstructs the image, yielding \( X_c \), we retains the original color but discards structural details. The process can be described as:
\begin{equation}
\mathcal{F}\left(\hat{x}\right)=\mathcal{A}\left(x\right) \times e^{\Phi}, \quad x_c = \mathcal{F}^{-1}\left(\mathcal{F}\left(\hat{x}\right)\right).
\end{equation}

\textbf{Texture Feature Extractor.} We emphasize the repetitive patterns and surface properties of the image while filtering out shape and color details. To remove color information, we first convert \( X \) from an RGB image to a grayscale image. Then, to preserve both local and global texture information and avoid excessive similarity in the texture, we adaptively crop the grayscale image into several square regions. Each square region is compared with all others for similarity using the Gray-Level Co-occurrence Matrix (GLCM). If the similarity between any two regions exceeds a threshold between 0.1 and 0.3, the region is stored independently. This process continues until four distinct square regions are collected. Once the regions are gathered, they are combined into a texture representation image \( X_t \), which captures the essential texture features.
\textbf{Shape Feature Extractor.} To highlight geometric shapes prioritized by the HVS, we apply entity segmentation followed by a pre-trained CNN and GradCAM \cite{chattopadhay2018grad} to extract shape-focused features, excluding color and texture. After obtaining separated feature samples $X_c$, $X_t$, and $X_s$, a self-attention classifier $f_a$ is trained to adaptively weight and aggregate them. Using the learned attention, $f_a$ dynamically combines feature modalities for task-relevant recognition.
\begin{equation}
P(Y \mid X) = f_a\left(f\left(X_c, X_t, X_s\right)\right).
\end{equation}

We perform causal interventions on separated feature samples, assuming that domain shift arises mainly from external attributes like brightness, geometry, and color, rather than intrinsic semantic changes. These external attributes are treated as confounders affecting the image features, and we design data transformation methods to adjust them. This approach improves model robustness to domain variations by mitigating distribution differences between source and target domains. We define confounding factors for each feature sample as \( H = \{ h_1^c, h_2^c, \dots, h_k^c \} \), where \( c \) corresponds to color, and each \( h_k^c \) is linked to a transformation function that modifies the respective external attribute to reduce confounding effects. For example, the transformation for the color factor \( C \) can be defined as:
\begin{equation}
\begin{aligned}
& x_{i}^{\hat{c}} = G_{h_k^c}\left(\cdots G_{h_2^c}\left(G_{h_1^c}\left(x_{i}^c ; \theta_{h_1^c}\right) ; \theta_{h_2^c}\right) \cdots ; \theta_{h_k^c}\right), \\
& \theta_{h_i^c} \in [u_k^{min}, u_k^{max}], \quad i \in [1, k],
\end{aligned}
\label{eqn4}
\end{equation}
where \( \theta_{h_i^c} \) represents the \( i_{th} \) data transformation function, bridging \( D_{so} \) and \( D_{tr} \) by generating a new data distribution. By applying \( \theta_{h_{k_{c}}^c} \), \( \theta_{h_{k_{t}}^t} \), and \( \theta_{h_{k_{s}}^s} \) across different visual attributes, we align and generalize distributions across domains. The weight parameters \( u_k^{min} \) and \( u_k^{max} \) control the transformation magnitude within a specific scale range. Through these transformations, we approximate the model’s generalization under the marginal distributions \( P(Y \vert X_c, Z_c) \) and \( P(Y \vert X_t, Z_t) \).

We calculate weighted averages based on sample proportions with different contextual prototypes to estimate the average causal effect. To isolate the influence of each causal factor, we apply transformations to \( X_c \) and \( X_t \) in \( V_c \) and \( V_t \), using varying types or degrees of distortion. To estimate the effects of \( Z_c \) and \( Z_t \), we use the back-door criterion from causal inference, applying do-calculus interventions to disentangle their effects. This allows independent assessment of \( P(Y \vert X_c) \) and \( P(Y \vert X_t) \), controlling for confounders and ensuring accurate causal learning. For instance, the back-door criterion in \( P(Y \vert X_c) \) is defined as:
\begin{equation}
\begin{aligned}
& P\left(Y \mid do\left(X_c\right)\right) = \sum_{Z_c} P\left(Y \mid X_c, Z_c\right) P\left(Z_c \mid X_c\right), \\
\end{aligned}
\end{equation}
The overall HSCM distribution can be formalized as:
\begin{equation}
\begin{aligned}
P(Y \mid do(X)) = & \, P\left(Y \mid X_c, Z_c\right) P\left(X_c\right) P\left(X \mid Z_c\right) + \\
& \, P\left(Y \mid X_t, Z_t\right) P\left(X_t\right) P\left(X \mid Z_t\right) + \\
& \, P\left(Y \mid X_c, Z_c\right) P\left(X_c\right) P\left(X \mid Z_t\right) + \\
& \, P\left(Y \mid X_t, Z_t\right) P\left(X_t\right) P\left(X \mid Z_c\right) + \\
& \, P(Y \mid S),
\end{aligned}
\end{equation}
where \( P(X_{c \backslash t}) \) be the marginal distribution of color and texture semantics at the dataset level, and \( P(X \mid Z_{c \backslash t}) \) represent the ratio of color and texture features for a given image. Assuming the independence of confounding factors in causal inference (i.e., \( P\left(X_c \mid Z_c\right) = P\left(X_t \mid Z_t\right) = 1 \) and \( P\left(X_c \mid Z_t\right) = P\left(X_t \mid Z_c\right) = 0 \)), the equation can be reformulated as follows:
\begin{equation}
\begin{aligned}
& P(Y \mid do(X)) := \\
& \quad P\left(Y \mid X_c, Z_c\right) + P\left(Y \mid X_t, Z_t\right) + P(Y \mid S).
\end{aligned}
\end{equation}

The confused category \( x_c \) for a given sample \( x_i^c \) can be obtained from its posterior probability as:
\begin{equation}
\begin{aligned}
& P\left(Y \mid X_c, Z_c\right) \\
& \quad = y_{i, h_k^c}^{\hat{c}} = \frac{1}{|\mathcal{M}|} \sum_{\theta_{h_k^c} \in \mathcal{M}} \mathcal{C}\left(f\left(G_{h_k^c}\left(x_i^c ; \theta_{h_k^c}^c\right)\right)\right).
\end{aligned}
\end{equation}

During model training, confounding samples that are easy to fit often contain spurious edges, which may appear relevant but do not reflect true causal relationships. These samples should have reduced weights, while more challenging samples with key causal information should be given higher weights. By adaptively adjusting the weights, the influence of genuine causal relationships is gradually emphasized, minimizing the impact of spurious edges. The adaptive weights are obtained as follows: 
\begin{equation}
\begin{aligned}
& a_{i,k}^c = P(Y \mid X) - P\left(Y \mid X_c, Z_c\right) = y_{i}^c - y_{i, h_k^c}^{\hat{c}}, \\
& w^c_i = \operatorname{softmax}\left(W_c\left(a_{i,1}^c\right), W_c\left(a_{i,2}^c\right), \cdots, W_c\left(a_{i,k}^c\right)\right),
\end{aligned}
\label{eqn5}
\end{equation}
where \( w_{i,k}^c \) represents the \( k \)-th weight-aware mapping in \( w^c_i \), and \( W(\cdot) \) denotes the effect-to-weight network. Generally, a larger value of \( a_{i,k}^c \) indicates a stronger causality of the confounding factor combinations. The loss related to the color confounding factor is calculated as follows:
\begin{equation}
\begin{aligned}
& \mathcal{L}_c = \mathcal{L}_{cro}^c + \frac{1}{N_{so}} \sum_i \mathcal{L}_{l2} \left(f\left(x_i^{so}\right),\sum_k w_{i,k}^c M_k^c\left(f\left(x_i^{\hat{c}}\right)\right)\right)+ \\
& \quad \frac{1}{N_{so}} \frac{1}{K} \sum_i \sum_k \mathcal{L}_{l2} \left(f\left(x_i^{so}\right),M_k^c\left(f\left(x_{i,k}^c\right)\right)\right) + \\
& \quad \frac{1}{N_{so}} \sum_i \mathcal{L}_{cro}\left(f_a\left(\sum_k w_{i,k}^c M_k^c\left(f\left(x_i^{\hat{c}}\right)\right), y_i^{so}\right)\right) + \\
& \quad \frac{1}{N_{so}} \frac{1}{K} \sum_i \sum_k \mathcal{L}_{cro}\left(f_a\left(M_k^c\left(f\left(x_{i,k}^c\right)\right), y_i^{so}\right)\right) + \\
& \quad \frac{1}{N_{so}} \frac{1}{K} \sum_i \sum_k \mathcal{L}_{cro}\left(f_a\left(M_k^c\left(f\left(x_{i,k}^c\right)\right), (M_k^t\left(f\left(x_{i,k}^t\right)\right)\right)\right),
\end{aligned}
\end{equation}
where \( \mathcal{L}_{cro} \) and \( \mathcal{L}_{l2} \) denote the cross-entropy and L2 losses, respectively. \( N_{so} \) is the total number of images in \( D_{so} \), and \( M_k^c \) is the weight-aware mapping that assigns importance weights to transformed samples. The second and fourth terms measure the feature distance and output logic between source and rescore-balanced samples after transformation, while the third and fifth terms assess independent data transformation. The sixth term encourages \( x_c \) and \( x_t \) samples to align with the source samples' categories. Thus, the overall loss \( L \) is formulated as:
\begin{equation}
\mathcal{L}=\mathcal{L}_c+\mathcal{L}_t+\mathcal{L}_{cro}^s .
\end{equation}

\textbf{Confounding Factors Optimization} Using a fixed number of confounding factors throughout network training, as shown in Eq. \ref{eqn4}, is suboptimal because confounders may distort causal effects by correlating with both independent and dependent variables. This limits the model's adaptability, hindering its ability to effectively control these latent factors and affecting causal inference accuracy. To address this, we evaluate \( a_{i,k}^c \) (Eq. \ref{eqn5}) at each iteration during training, using it to select appropriate data transformations \( G_{h_k^c} \) (Eq. \ref{eqn4}). The process can be formulated as follows:
\begin{equation}
\textcolor[RGB]{0,0,0}{
\begin{aligned}
\mathcal{A} & =\left\{a_{i,1}^c (\tau), a_{i,2}^c (\tau), \cdots, a_{i,k}^c (\tau)\right\},
\end{aligned}}
\end{equation}
where \( \mathcal{A} \) be a set containing \( k \) elements, representing the influence of the \( k_{th} \) data transformation \( G \) at the \( \tau_{th} \) epoch. We first sort \( \mathcal{A} \) in descending order to obtain \( \mathcal{A}_{sorted} \). Based on this, we adjust the data transformations: if the loss function decreases significantly, we reduce the most influential transformation; if the decrease is small, we randomly select a transformation from the reduced set. This process is formally defined as follows:
\begin{equation}
\begin{aligned}
G_{\text {remove}} &= \max_{} \operatorname{impact}\left(G_{i,k}\right) = \mathcal{A}_{sorted}[:1], \\
\Delta \mathcal{L} &= \mathcal{L}(\tau-1) - \mathcal{L}(\tau), \\
G_{\text {remain}} &= 
\begin{cases}
G \backslash\left\{G_{\text {remove}}\right\}, & \text{if } \Delta L > \epsilon_L  \\
G \backslash\{\operatorname{random}(G)\}, & \text{if } \Delta L \leq \epsilon_L
\end{cases}
\end{aligned}
\end{equation}
where \( \Delta L \) represents the change in the loss function between iterations, and \( \epsilon_L \) is a threshold for adjusting the number of data transformations based on the loss decrease. As a hyperparameter, \( \epsilon_L \) governs weight adjustments. During early training, a larger \( \epsilon_L \) allows significant weight changes to capture global features. As training progresses and stabilizes, a smaller \( \epsilon_L \) increases sensitivity to subtle loss changes, enabling finer adjustments while preventing overfitting or underfitting. This process can be implemented through a straightforward strategy as follows:
\begin{equation}
\epsilon_L(\tau)=\epsilon_{L, 0} \cdot\left(\frac{\tau_{all}-\tau}{\tau_{all}}\right),
\end{equation}
where \( \epsilon_L, 0 \) is the initial value of \( \epsilon_L \), \( \tau \) is the current training iteration, and \( \tau_{all} \) is the maximum number of training epochs. As training progresses, \( \epsilon_L \) will gradually decrease.


\section{Experiments}
\subsection{Datasets and Setup}

\noindent\textbf{Datasets.} Our experiments involve five digit datasets and three real-world scenarios datasets. The digit datasets \cite{zhou2020deep} contain one random-selected pairs with MNIST \cite{lecun1998gradient} from four digit datasets: SVHN \cite{netzer2011reading}, SYN \cite{dou2019domain}, MNIST-M \cite{janzing2010causal}, and USPS \cite{hull1994database} (e.g., MNIST \(\rightarrow \) SVHN and MNIST \(\rightarrow \) USPS). The digit datasets are benchmark datasets used for handwritten digit recognition, differing in their sources, image styles, and complexity. Specifically, MNIST and USPS contain handwritten digits, SVHN comes from street view house numbers, SYN consists of synthetic digits, and MNIST-M introduces domain shift through variations in color and background. The real-world scenarios datasets is more challenging, specifically in terms of CIFAR-10 \cite{krizhevsky2009learning} \(\rightarrow \) CIFAR-10-C \cite{hendrycks2019benchmarking}, PACS \cite{li2017deeper}, Office-Home \cite{venkateswara2017deep}. Amongthem, CIFAR-10 is a classic image classification dataset with 60,000 32x32 color images across 10 categories. CIFAR-10-C extends CIFAR-10 with image corruption factors (e.g., noise, blur, color distortion) to test model robustness on low-quality images. PACS is a cross-domain dataset with four domains: photos, artworks, cartoons, and sketches, used for domain adaptation and generalization. Office-Home, a multi-domain dataset, includes art, clipart, products, and real-world images, and is designed for domain adaptation and cross-domain learning. These datasets assess model performance under various conditions, from standard classification (MNIST, CIFAR-10) to robustness (SVHN, SYN, MNIST-M, USPS, and CIFAR-10-C) and cross-domain learning (PACS, Office-Home).

\noindent\textbf{Implementation Details.} We define 16 data transformations to simulate the effects of different confounding factors: \( Z_c \) (Brightness, Contrast, Color, Sharpness, AutoContrast, Invert, Equalize, Solarize, SolarizeAdd, Posterize), \( Z_t \) (NoiseSalt, NoiseGaussian), and \( Z_s \) (ShearX, Shear-Y, Rotate, Flip). These data transformations are designed to capture the influence of various visual attributes on the data, thereby allowing us to analyze the impact of confounding factors across different domains. The hardware configuration includes an Intel Core i9-13900K CPU running at 5.50 GHz, 64GB of RAM in a dual-channel configuration, a 2TB ROM, and five RTX 3090 GPUs. This system runs on Ubuntu 20.04.5 LTS, with Python 3.9.12 and PyTorch 2.5.1 installed.

\begin{table*}[t]
  \centering
  \caption{Leave-one-domain-in results on Digits with ConvNet.}
  \resizebox{0.6\textwidth}{!}{
    \begin{tabular}{l|ccccc}
    \midrule
    Methods & \multicolumn{1}{c|}{SVHN} & \multicolumn{1}{c|}{SYN} & \multicolumn{1}{c|}{MNIST-M} & \multicolumn{1}{c|}{USPS} & Avg. \\
    \midrule
    ERM   & 27.83  & 39.65  & 52.72  & 76.94  & 49.29  \\
    CCSA  & 25.89  & 37.31  & 49.29  & 83.72  & 49.05  \\
    D-SNE  & 26.22  & 37.83  & 50.98  & \textbf{93.16}  & 52.05  \\
    JIGEN  & 33.80  & 43.79  & 57.80  & 77.15  & 53.14  \\
    GUD   & 35.51  & 45.32  & 60.41  & 77.26  & 54.62  \\
    M-ADA  & 42.55  & 48.95  & 67.94  & 78.53  & 59.49  \\
    ME-ADA  & 42.56  & 50.39  & 63.27  & 81.04  & 59.32  \\
    PDEN  & 62.21  & 69.39  & 82.20  & 85.26  & 74.77  \\
    L2D   & 62.86  & 63.72  & \textbf{87.30}  & 83.97  & 74.46  \\
    AA    & 45.23  & 64.52  & 60.53  & 80.62  & 62.72  \\
    RA    & 54.77  & 59.60  & 74.05  & 77.33  & 66.44  \\
    RSDA  & 47.40  & 62.00  & 81.50  & 83.10  & 68.50  \\
    RSDA+ASR  & 52.80  & 64.50  & 80.80  & 82.40  & 70.10  \\
    META  & \underline{69.94}  & \underline{78.47}  & 78.34  & 88.54  & \underline{78.82}  \\
    \midrule
    Ours  & \textbf{73.21} & \textbf{83.90} & \underline{78.91} & \underline{89.24} & \textbf{81.32} \\
    \bottomrule
    \end{tabular}
    }
  \label{tab1}
\end{table*}

\begin{table*}[t]
  \centering
  \caption{Leave-one-domain-in results on CIFAR10-C with WRN.}
    \resizebox{0.6\textwidth}{!}{
    \begin{tabular}{l|cccccc}
    \midrule
    Methods & \multicolumn{1}{c|}{L1} & \multicolumn{1}{c|}{L2} & \multicolumn{1}{c|}{L3} & \multicolumn{1}{c|}{L4} & \multicolumn{1}{c|}{L5} & Avg. \\
    \midrule
    ERM   & 87.80  & 81.50  & 75.50  & 68.20  & 56.10  & 73.82  \\
    GUD   & 88.30  & 83.50  & 77.60  & 70.60  & 58.30  & 75.66  \\
    M-ADA  & 90.50  & 86.80  & 82.50  & 76.40  & 65.60  & 80.36  \\
    PDEN  & 90.62  & 88.91  & 87.03  & 83.71  & 77.47  & 85.55  \\
    AA    & 91.42  & 87.88  & 84.10  & 78.46  & 71.13  & 82.60  \\
    RA    & 91.74  & 88.89  & 85.82  & 81.03  & 74.93  & 84.48  \\
    META  & \underline{92.38}  & \underline{91.22}  & \underline{89.88}  & \underline{87.73}  & \underline{84.52}  & \underline{89.15}  \\
    \midrule
    Ours  &   \textbf{93.28} & \textbf{92.64} & \textbf{91.36} & \textbf{93.33} & \textbf{94.44} & \textbf{93.01}  \\
    \bottomrule
    \end{tabular}
    }
  \label{tab2}
\end{table*}

\begin{table*}[t]
  \centering
  \caption{Leave-one-domain-in results on PACS with ResNet-18.}
    \resizebox{0.6\textwidth}{!}{
    \begin{tabular}{l|ccccc}
    \midrule
    Methods & \multicolumn{1}{c|}{Art} & \multicolumn{1}{c|}{Cartoon} & \multicolumn{1}{c|}{Sketch} & \multicolumn{1}{c|}{Photo} & Avg. \\
    \midrule
    ERM   & 70.90  & 76.50  & 53.10  & 42.20  & 60.70  \\
    RSC   & 73.40  & 75.90  & 56.20  & 41.60  & 61.80  \\
    RSC+ASR  & 76.70  & 79.30  & 61.60  & 54.60  & 68.10  \\
    META  & \underline{77.13}  & \underline{80.14}  & \underline{62.55}  & \underline{59.60}  & \underline{69.86}  \\
    \midrule
    Ours  & \textbf{77.98} & \textbf{81.22} & \textbf{66.90} & \textbf{61.00} & \textbf{71.78} \\
    \bottomrule
    \end{tabular}
  \label{tab3}
  }
\end{table*}

\subsection{Results on Single Domain Generalization}

As shown in Tables \ref{tab1}, \ref{tab2}, and \ref{tab3}, our proposed method achieves the highest average accuracy across all compared methods in domain generalization tasks, surpassing all alternatives in performance. These results are drawn from experiments conducted on both the leave-one-domain-out digit datasets and real-world scenario datasets, each presenting unique challenges. More specifically, the results highlight that our approach can handle the inherent complexity and variability in both synthetic and real-world environments, managing to reduce the impact of domain-specific discrepancies effectively. While traditional methods often struggle to maintain performance under domain shift, our method retains high accuracy across all tested domains. This indicates its robustness in mitigating the effects of changes in visual attributes, environmental conditions, and other potential confounding factors. Overall, the performance of our method in these experiments strongly suggests that it is a promising solution for real-world applications where domain shift and generalization are key concerns. By effectively addressing these challenges, our approach sets a new benchmark for domain generalization, showcasing its potential in diverse real-world settings where conventional methods often fall short.


\begin{table*}[t]
  \centering
  \caption{Leave-one-domain-out results on PACS with ResNet-18.}
    \resizebox{0.6\textwidth}{!}{
    \begin{tabular}{l|ccccc}
    \midrule
    Methods & \multicolumn{1}{c|}{Art} & \multicolumn{1}{c|}{Cartoon} & \multicolumn{1}{c|}{Photo} & \multicolumn{1}{c|}{Sketch} & Avg. \\
    \midrule
    MetaReg  & 83.70  & 77.20  & 95.50  & 70.30  & 81.70  \\
    JiGen  & 79.42  & 75.25  & 96.03  & 71.35  & 80.51  \\
    GUD  & 78.32  & 77.65  & 95.61  & 74.21  & 81.44  \\
    Epi-FCR  & 82.10  & 77.00  & 93.90  & 73.00  & 81.50  \\
    DMG  & 76.90  & 80.38  & 93.55  & 75.21  & 81.46  \\
    DDAIG  & 84.20  & 78.10  & 95.30  & 74.70  & 83.10  \\
    CSD   & 78.90  & 75.80  & 94.10  & 76.70  & 81.40  \\
    MASF  & 80.29  & 77.17  & 94.99  & 71.69  & 81.04  \\
    L2A-OT  & 83.30  & 78.20  & 96.20  & 73.60  & 82.80  \\
    EISNet  & 81.89  & 76.44  & 95.93  & 74.33  & 82.15  \\
    MatchDG  & 81.32  & 80.70  & 95.63  & 79.72  & 84.56  \\
    ME-ADA  & 78.61  & 78.65  & 95.57  & 75.59  & 82.10  \\
    MMLD  & 81.28  & 77.16  & 96.09  & 72.29  & 81.83  \\
    L2D  & 81.44  & 79.56  & 95.51  & 80.58  & 84.27  \\
    RSC   & 83.43  & 80.31  & 95.99  & 80.85  & 85.15  \\
    FACT  & 85.90  & 79.35  & \underline{96.61}  & 80.88  & 85.69  \\
    CIRL  & \underline{86.08}  & 80.59  & 95.93  & 82.68  & 86.32  \\
    META  & 85.30  & \underline{80.93}  & 96.53  & \underline{85.24}  & \underline{87.00}  \\
    \midrule
    Ours  & \textbf{87.21} & \textbf{81.36} & \textbf{97.13} & \textbf{85.26} & \textbf{87.74} \\
    \bottomrule
    \end{tabular}
  \label{tab4}
  }
\end{table*}

\begin{table*}[t]
  \centering
  \caption{Leave-one-domain-out results on Office-Home with ResNet-18.}
    \resizebox{0.6\textwidth}{!}{
    \begin{tabular}{l|ccccc}
    \midrule
    Methods & \multicolumn{1}{c|}{Art} & \multicolumn{1}{c|}{Clipart} & \multicolumn{1}{c|}{Product} & \multicolumn{1}{c|}{Real} & Avg. \\
    \midrule
    CCSA  & 59.90  & 49.90  & 74.10  & 75.70  & 64.90  \\
    MMD-AAE & 56.50  & 47.30  & 72.10  & 74.80  & 62.70  \\
    CrossGrad  & 58.40  & 490.40  & 73.90  & 75.80  & 64.40  \\
    DDAIG  & 59.20  & 52.30  & 74.60  & 76.00  & 65.50  \\
    L2A-OT  & 60.60  & 50.10  & 74.80  & \underline{77.00} & 65.60  \\
    JiGen  & 53.04  & 47.51  & 71.47  & 72.79  & 61.20  \\
    RSC   & 58.42  & 47.90  & 71.63  & 73.54  & 63.12  \\
    FACT  & 60.34  & 54.85  & 74.48  & 76.55  & 66.56  \\
    CIRL  & \underline{61.48} & \underline{55.28} & \underline{75.06} & 76.64  & \underline{67.12}  \\
    \midrule
    Ours  & \textbf{62.22} & \textbf{56.13} & \textbf{75.53} & \textbf{77.27} & \textbf{67.79}  \\
    \bottomrule
    \end{tabular}
  \label{tab5}
  }
\end{table*}

\begin{table*}[t]
  \centering
  \caption{Leave-one-domain-out results on PACS with ResNet-50.}
    \resizebox{0.6\textwidth}{!}{
    \begin{tabular}{l|ccccc}
    \midrule
    Methods & \multicolumn{1}{c|}{Art} & \multicolumn{1}{c|}{Cartoon} & \multicolumn{1}{c|}{Photo} & \multicolumn{1}{c|}{Sketch} & Avg. \\
    \midrule
    DDAIG  & 84.94  & 76.98  & 97.64  & 76.75  & 84.08  \\
    MetaReg  & 87.20  & 79.20  & 97.60  & 70.30  & 83.60  \\
    MASF  & 82.89  & 80.49  & 95.01  & 72.29  & 82.67  \\
    EISNet  & 86.64  & 81.53  & 97.11  & 78.07  & 85.84  \\
    MatchDG  & 85.61  & 82.12  & 97.94  & 78.76  & 86.11  \\
    FACT  & \underline{90.89}  & 83.65  & \underline{97.98}  & 86.17  & 89.62  \\
    CIRL  & 90.67  & \underline{84.30}  & 97.84  & \underline{87.68}  & \underline{90.12}  \\
    \midrule
    Ours  & \textbf{91.65} & \textbf{85.75} & \textbf{98.32} & \textbf{88.52} & \textbf{91.06} \\
    \bottomrule
    \end{tabular}
  \label{tab6}
  }
\end{table*}

\subsection{Results on Multiple Domain Generalization}

In this experiment, we extend our method to the multi-source domain setting by treating multiple source domains as a unified source, allowing for simultaneous utilization of cross-domain information while ensuring strong generalization. Using the leave-one-domain-out protocol, a standard benchmark for multi-source domain generalization, we evaluate our approach on three challenging real-world datasets. As demonstrated in Tables \ref{tab4}, \ref{tab5}, and \ref{tab6}, our method consistently achieves the highest average performance, showcasing its resilience under significant domain shifts and variability. On the PACS dataset, our method surpasses the advance methods by 0.74\% and 0.94\%. On the Office-Home dataset, which features more categories and higher visual diversity compared to PACS, our method surpasses the CIRL model by 0.67\%. This result highlights our approach's robustness and its capacity to effectively capture complexities in multi-source domain settings, excelling even in large-scale, diverse scenarios.


\subsection{Interpretabilities on Model Decisions}

\begin{figure*}[t!]
    \centering
    \includegraphics[width=0.7\columnwidth]{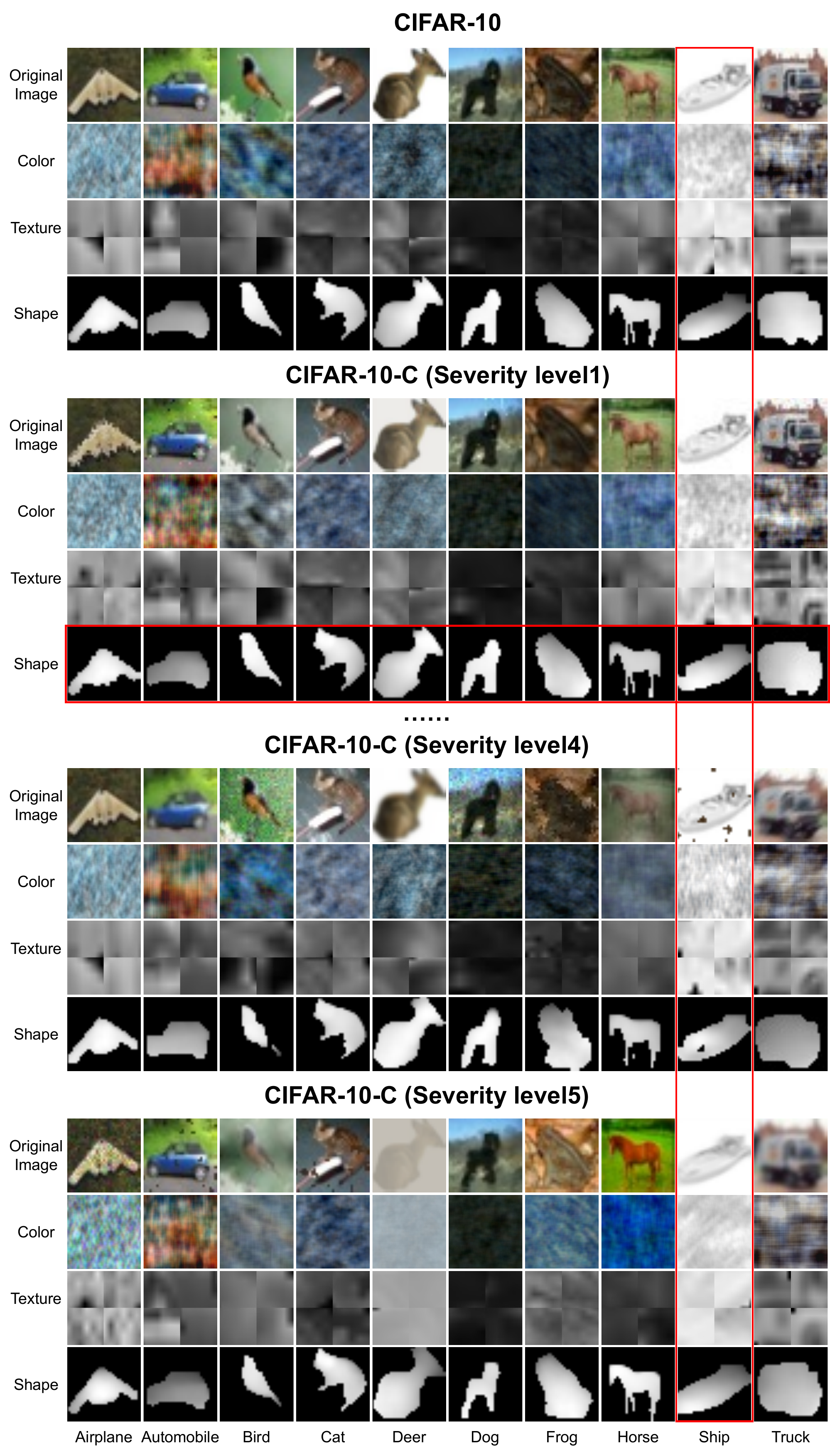}
    \setlength{\abovecaptionskip}{0.00cm}
	\setlength{\belowcaptionskip}{0.0cm}
    \vspace{-0.00em}
    \caption
	{
		 Interpretability visualization for CIFAR-10 and CIFAR-10-C classes.
    }
    \label{fig4}
\end{figure*}

\begin{figure*}[t!]
    \centering
    \includegraphics[width=1.0\columnwidth]{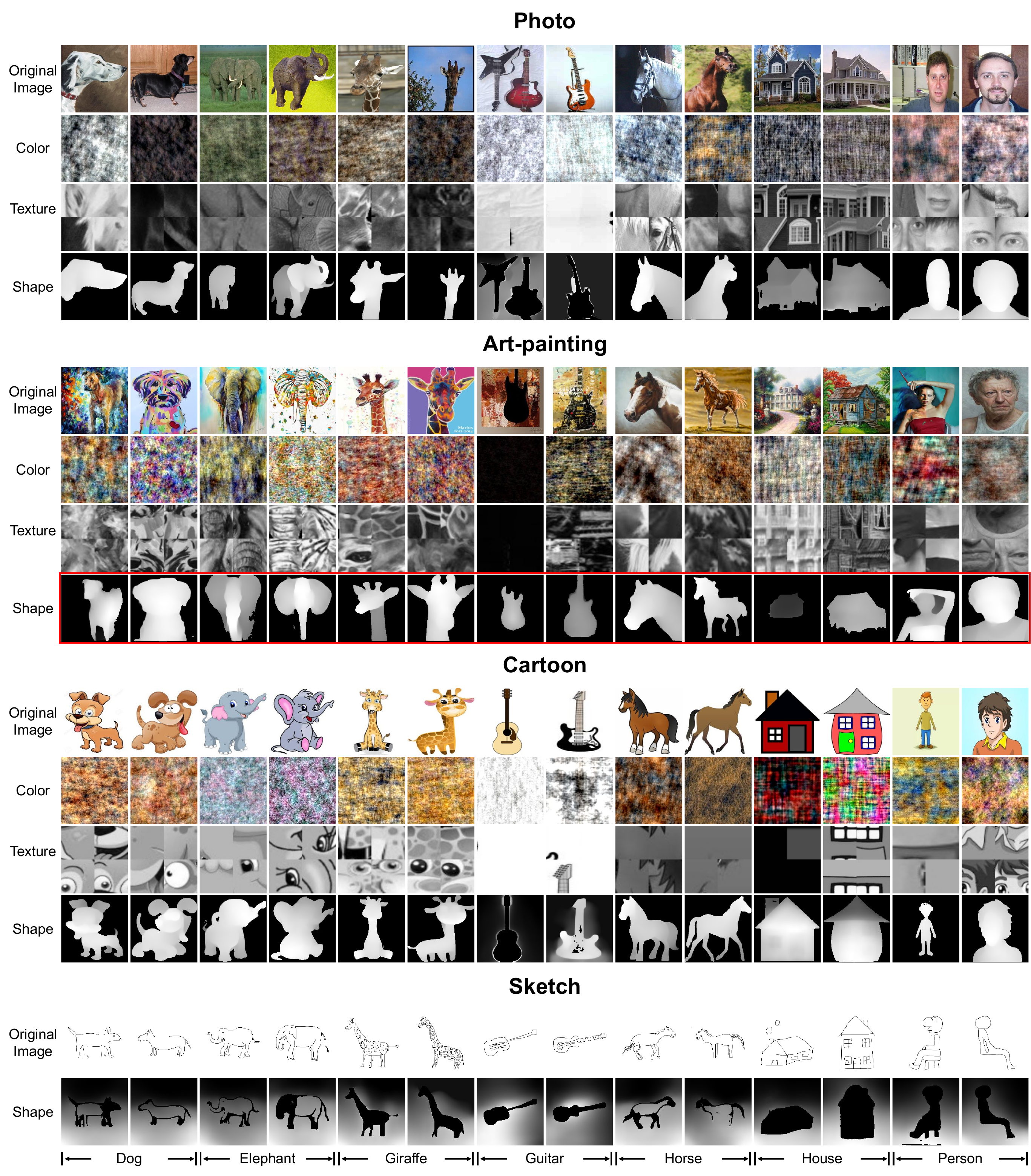}
    \setlength{\abovecaptionskip}{0.00cm}
	\setlength{\belowcaptionskip}{0.0cm}
    \vspace{-0.00em}
    \caption
	{
		 Interpretability visualization for PACS classes.
    }
    \label{fig5}
    \vspace{-1.00em}
\end{figure*}

\begin{figure*}[t!]
    \centering
    \includegraphics[width=1.0\columnwidth]{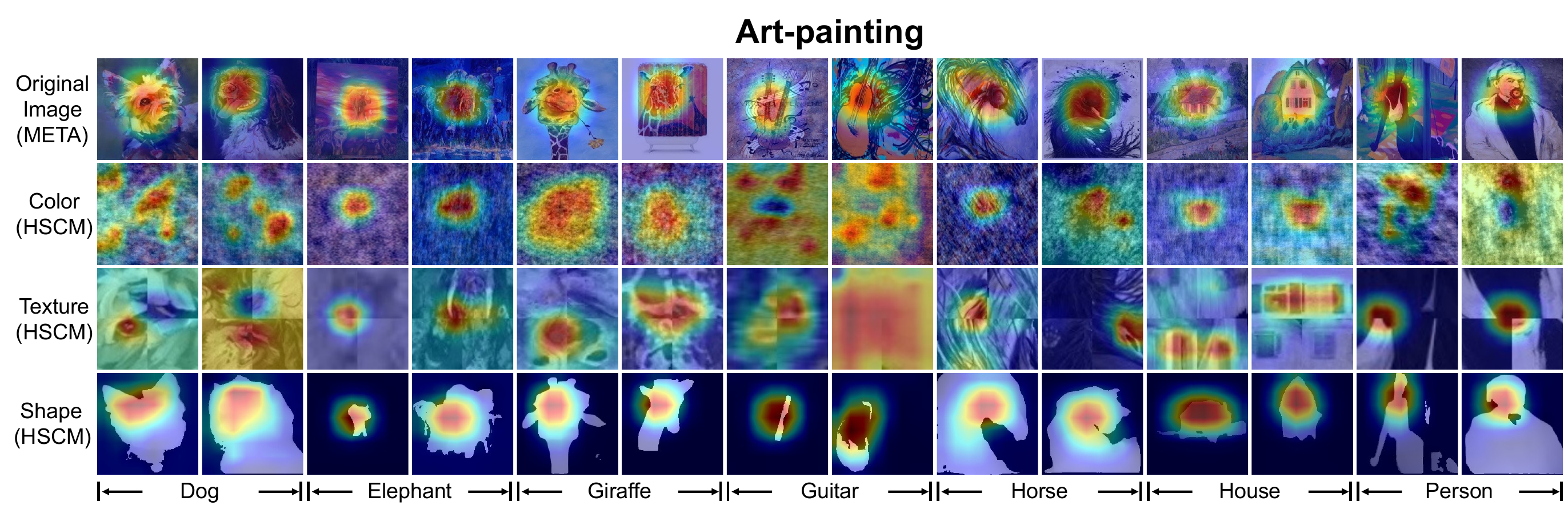}
    \setlength{\abovecaptionskip}{0.00cm}
	\setlength{\belowcaptionskip}{0.0cm}
    \vspace{-0.00em}
    \caption
	{
		 CAMs \cite{chattopadhay2018grad} are supervised by META and the HSCM framework on PACS dataset, which focuses on different visual attributes, namely color, texture, and shape. The first row shows the original images and their corresponding CAMs supervised by META, while the second, third, and fourth rows display the CAMs for HSCM's color, texture, and shape features, respectively.
    }
    \label{fig6}
\end{figure*}

\begin{figure*}[t!]
    \centering
    \includegraphics[width=1.0\columnwidth]{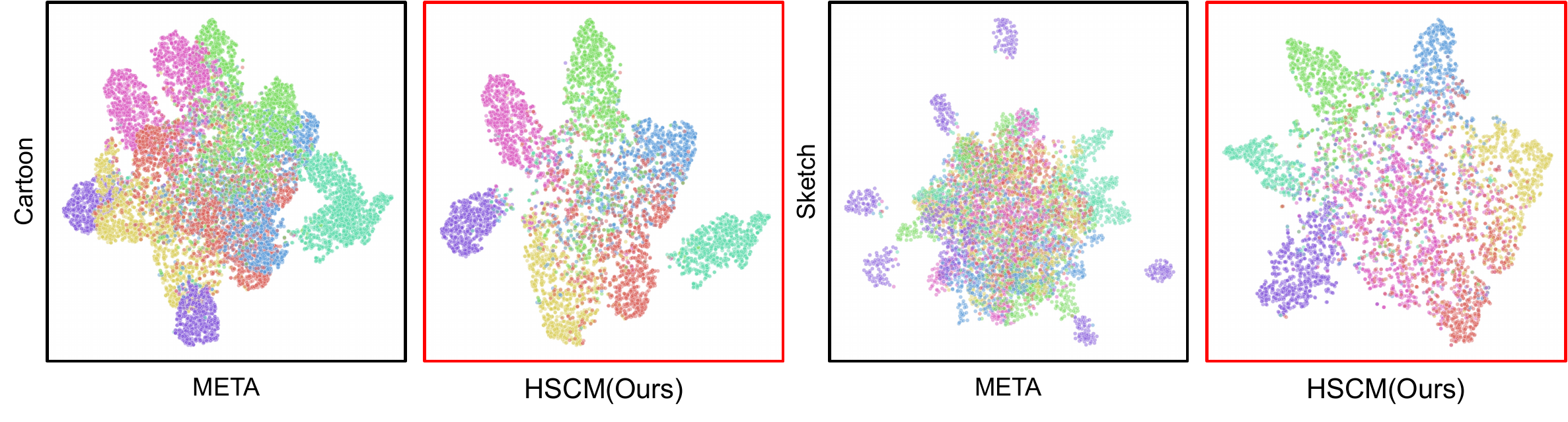}
    \setlength{\abovecaptionskip}{0.00cm}
	\setlength{\belowcaptionskip}{0.0cm}
    \vspace{-0.00em}
    \caption
	{
		 The t-SNE \cite{hinton2002stochastic} visualizations show the feature distributions learned by META (black border) and HSCM (red border). These features are from a random selection of 7 categories from the PACS: dog, elephant, giraffe, guitar, horse, house, and person.
    }
    \label{fig7}
    \vspace{-0.00em}
\end{figure*}

We explain the interpretability of HSCM from three perspectives: at the model input, we visualize the separation of color, texture, and shape features, aligning with human visual perception. During model processing, we use Class Activation Mapping (CAM) \cite{chattopadhay2018grad} to show how these features contribute to model decisions. And at the model output, t-SNE \cite{hinton2002stochastic} visualizations demonstrate how HSCM effectively classifies objects by leveraging these attributes.

\textbf{Model input.} Fig. \ref{fig4} presents the interpretability visualization of the HSCM framework on CIFAR-10 and CIFAR-10-C datasets under varying corruption severities. We randomly selected ten samples from each CIFAR-10 class for a comprehensive analysis. Each row corresponds to different feature components, such as color, texture, and shape, which extracted by HSCM. As corruption severity increases (from clean CIFAR-10 to CIFAR-10-C with severity levels 1, 4, and 5), the color and texture components exhibit progressive degradation. Specifically, the color component becomes more distorted and shifts from a clear representation to random noise under high-severity corruptions. The texture component shows increasing levels of blurring and structural disruption, especially under higher severity, making it challenging to capture fine-grained details. In contrast, the shape component remains stable across all corruption levels, consistently capturing the object's geometric structure. This stability highlights the HSCM framework’s ability to preserve critical object shapes while isolating noise-sensitive features like color and texture. Notably, the shape representation retains clear boundaries even under the most severe corruptions (severity level 5), emphasizing its robustness in maintaining object structure.

Fig. \ref{fig5} illustrates the interpretability of the HSCM framework on the PACS dataset, showcasing its ability to disentangle color, texture, and shape features across various domains: Photo, Art-painting, Cartoon, and Sketch. We randomly selected two images from each category for visualization to highlight how HSCM separates these visual factors. In the Photo domain, the color component captures realistic patterns, the texture component identifies surface details, and the shape component isolates object outlines. In Art-painting, the color component captures vibrant hues, the texture component emphasizes artistic brushstrokes, and the shape component isolates object outlines, albeit less clearly than in photos. In Cartoon, HSCM extracts exaggerated colors, stylized textures, and simplified shapes, highlighting the model’s ability to handle stylized visual features. In the Sketch domain, where color and texture are minimal, the color and texture components produce black images, while the shape component effectively captures object outlines. This demonstrates HSCM’s robustness in domains with limited color and texture information.

\textbf{Model process.} Fig. \ref{fig6} shows CAMs supervised by META and the HSCM framework. In the HSCM visualization, the color component highlights areas of the image with strong color patterns, emphasizing how color influences the model's decision for objects like Dog, Elephant, and Guitar. The texture component captures finer surface details and patterns, contributing to the decision-making process in cases like the Horse and House images. The shape component isolates the structural outlines of objects, showing how critical shape information is for classification, as seen in objects like Giraffe and Person. This visualization illustrates HSCM's ability to focus on different visual aspects of an image, reflecting a more nuanced understanding of object features that aligns closely with human visual perception. By separately analyzing these features, HSCM enhances the model's interpretability and robustness, enabling it to make more informed decisions based on distinct, meaningful visual attributes.

\textbf{Model output.} Fig. \ref{fig7} compares the META model with the proposed HSCM framework across cartoon and sketch domains. Each point represents an image in the dataset, with different colors corresponding to distinct classes. In the Cartoon and Sketch domains, the META model shows significant class overlap, where the embeddings are more scattered, indicating that the model struggles to preserve class separability in these domains. The lack of clear separation between classes highlights the model's difficulty in handling domain shifts effectively. In contrast, the HSCM (Ours) framework exhibits much clearer class separation across both domains. The embeddings are more distinctly grouped, with minimal overlap between classes. This improved separation suggests that HSCM’s ability to disentangle key features, such as color, texture, and shape, allows it to better manage domain variability, improving class separability and generalization across domains. These results demonstrate that HSCM enhances the model's ability to learn more robust and domain-invariant representations, leading to better generalization performance, particularly in transfer tasks with visual differences.

\subsection{Ablation Studies}

As shown in Table \ref{tab7}, the ablation study on the PACS dataset clearly demonstrates the effectiveness of the individual components within the HSCM framework. Beginning with a base accuracy of 54.97\%, the introduction of T substantially boosts performance to 71.17\%. When C is integrated on top of the base model, accuracy increases to 61.98\%. Finally, the complete HSCM framework—combining T, C, and O, achieves the highest accuracy of 71.78\%. These results highlight the complementary roles of the three components. Specifically, T enhances the model’s capacity to disentangle domain-specific features, thereby reducing feature entanglement across domains. C addresses the influence of spurious correlations, ensuring that the model captures causal rather than coincidental patterns. Meanwhile, O adaptively reweights samples, assigning greater importance to informative data points and reducing the impact of outliers or redundant samples. Taken together, these mechanisms substantially improve both generalization and causal discovery, enabling the model to remain robust when applied to heterogeneous domains. The ablation study therefore validates not only the contribution of each individual component but also the synergy achieved when they are combined within the HSCM framework.

\begin{table*}[t]
  \centering
  \caption{Ablation study of on PACS dataset with ResNet-18. T, C, and O represent Data Transformations, Causal Intervention, and Confounding Factors Optimization, respectively.}
  \resizebox{0.70\textwidth}{!}{
    \begin{tabular}{l|ccc|cccc|c}
    \midrule
    Methods & \multicolumn{1}{c|}{T} & \multicolumn{1}{c|}{C} & O     & \multicolumn{1}{c|}{Art} & \multicolumn{1}{c|}{Cartoon} & \multicolumn{1}{c|}{Photo} & Sketch & Avg. \\
    \midrule
    Base  &       &       &       &    71.26    &   67.74     &   43.97     &   36.99     &  54.97 \\
    Variant 1 &   $\checkmark$    &       &       &   74.58    &   72.10    &   52.12    &    49.12   &  61.98\\
    Variant 2 &   $\checkmark$    &    $\checkmark$   &       &    76.92   &   78.65    &   60.13    &   57.45    &  68.29\\
    \midrule
    Ours  &   $\checkmark$  &  $\checkmark$ &  $\checkmark$ &   \textbf{77.98} & \textbf{81.22} & \textbf{66.90} & \textbf{61.00} & \textbf{71.78}  \\
    \bottomrule
    \end{tabular}
    }
  \label{tab7}
  \vspace{-0.00em}
\end{table*}

\section{Discussion and Conclusion}

The HSCM framework, inspired by the human visual system, shows promising results in domain generalization and causal representation learning by disentangling images into color, texture, and shape attributes and dynamically reweighting samples to mitigate spurious correlations. Theoretically, it ensures consistency and identifiability, converging to the true causal structure, while empirically it significantly enhances robustness, cross-domain generalization, and causal discovery stability. Nonetheless, current reliance on predefined feature extractors may limit its ability to capture the complexity of dynamic or abstract visual domains, and its performance under extreme domain shifts could benefit from advanced adversarial training or self-supervised learning to better handle noise and outliers. Future efforts will focus on refining the model’s flexibility, robustness, and computational efficiency, further advancing the field of interpretable, causal-driven domain generalization.

\begin{appendices}









\clearpage
\appendix

\section{Proof for Theorem1}
\textbf{Proof.} To analyze the expected cross-entropy loss of the classifier on an unseen test distribution, we decompose \( X \) into components related to \( Y \), specifically \( c \), \( t \), and \( s \). This decomposition reveals that the noise in \( X \) originates from distinct latent factors. Consequently, \( Y \) is conditionally independent of \( c \) and \( t \) given \( X \), i.e., \( Y \perp_{H} (c, t) \mid X \). Thus, Eq. 4 can be further rephrased as:
\begin{equation}
\begin{aligned}
L_{e_{tr}} & \left(p_{H}(Y \mid X)\right) = 
L_{e_{tr}}\left(p_{H}(Y)\right)+ \\&
\mathbb{E}_{p_{e_{tr}}(Y, c, t, s)}\left[\mathbb{E}_{p_{H}(X \mid Y, c, t, s)}\left[\log \frac{p_{H}(Y \mid c, t)}{p_{H}(Y \mid X, c, t)}\right]\right] \\
& =L_{e_{tr}}\left(p_{H}(Y)\right)+ \\&
\mathbb{E}_{p_{e_{tr}}(Y, c, t, s)}\left[\mathbb{E}_{p_{H}(X \mid Y, c, t, s)}\left[\log \frac{p_{H}(X \mid c, t)}{p_{H}(X \mid Y, c, t)}\right]\right] \\
& =L_{e_{tr}}\left(p_{H}(Y)\right)- \\&
\mathbb{E}_{p_{e_{tr}}(Y, c, t, s)} K L\left[p_{H}(X \mid Y, c, t) \| p_{H}(X \mid c, t)\right] .
\end{aligned}
\end{equation}

Thus, the cross-entropy loss of \( p_{H}(Y \mid X) \) in any environment \( e_{tr} \) is smaller than that of \( p_{H}(Y) \). The above description can be defined as:
\begin{equation}
L_{e_{tr}}\left(p_{H}(Y \mid X)\right)-L_{e_{tr}}\left(p_{H}(Y)\right) \leq 0.
\end{equation}

Therefore, the proof of Theorem 1 reduces to proving the equality \( p_{e_{so}}(Y \mid X) = p_{H}(Y \mid X) \). We apply the theorem of Bayes \cite{rish2001empirical, motiian2017unified} to express its conditional probability as formulated below:
\begin{equation}
\begin{aligned}
p_{e_{so}}(Y \mid X) & =p_{e_{so}}(Y \mid X, c, t, s) \\
& =p_{e_{so}}(Y) \frac{p_{e_{so}}(X \mid Y, c, t, s)}{\mathbb{E}_{p_{e_{so}}(Y \mid c, t, s)}\left[p_{e_{so}}(X \mid Y, c, t, s)\right]} \\
& =p_{H}(Y) \frac{p_{H}(X \mid Y, c, t, s)}{\mathbb{E}_{p_{H}(Y \mid c, t)}\left[p_{H}(X \mid Y, c, t, s)\right]} \\
& =p_{H}(Y \mid X, c, t, s) \\
& =p_{H}(Y \mid X).
\end{aligned}
\end{equation}

Thus, we have the minimax optimality of the classifier trained on the HSCM, as it encourages the model to capture features consistent across different environments. By addressing domain shift and focusing on invariant features, the classifier generalizes better to unseen data. In essence, the HSCM framework helps learn representations that are both discriminative and invariant, boosting the model's robustness and adaptability.

\section{Proof for Theorem2}
\textbf{Proof.} To learn a faithful DAG from a sample's joint distribution, our goal is to identify the optimal causal graph \( G \) by minimizing the negative log-likelihood loss, approximating the marginal likelihood of \( X \) conditioned on \( Y \) and \( G \), starting with the conditional log-likelihood \( \log p(X | Y, G) \).
\begin{equation}
\log p(X \mid Y, G) = \log \int p(X, c, t, s \mid Y, G) \, \text{d}c \, \text{d}t \, \text{d}s,
\end{equation}
where \( Z = \{c, t, s\} \) represents the latent variables. Here, \( s \) is a stable feature with respect to \( Y \), while \( c \) and \( t \) are environment-dependent noise factors. Since direct computation of this marginal likelihood is challenging, we introduce a variational distribution \( q(c, t, s | X, Y, G) \) to approximate the posterior. By applying Jensen’s inequality, we obtain the Evidence Lower Bound (ELBO) as follows:
\begin{equation}
\begin{aligned}
& \log p(X | Y, G) = \\
& \log \int q(c, t, s | X, Y, G) \frac{p(X, c, t, s | Y, G)}{q(c, t, s | X, Y, G)} \, \text{d}c \, \text{d}t \, \text{d}s \\
& \geq \mathbb{E}_{q(c, t, s | X, Y, G)} \left[ \log \frac{p(X, c, t, s | Y, G)}{q(c, t, s | X, Y, G)} \right] := \operatorname{ELBO}(X,Y,G).
\end{aligned}
\end{equation}

Expanding the above formula, the ELBO becomes:
\begin{equation}
\begin{aligned}
& \mathcal{L}(X, Y, G) = \\
& \mathbb{E}_{q(c, t, s | X, Y, G)} \left[ \log p(X | c, t, s, Y, G) \right] - \\ 
& D_{\text{KL}}\left( q(c, t, s | X, Y, G) \parallel p(c, t, s | Y, G) \right).
\end{aligned}
\label{eqn1}
\end{equation}

We introduce the adaptive weights \( W = (w_1, \ldots, w_i) \) for each sample to reweight the KL terms, prioritizing samples that are more informative. This allows the model to focus more on challenging and outlier samples. The reweighted ELBO is given by:
\begin{equation}
\begin{aligned}
& \mathcal{L}(X, Y, G) = \\
& \mathbb{E}_{q(c, t, s | X, Y, G)} \left[ \log p(X | c, t, s, Y, G) \right] - \\
& \sum_i w_i \cdot D_{\text{KL}}\left( q(c, t, s | X_i, Y, G) \parallel p(c, t, s | Y, G) \right).
\end{aligned}
\end{equation}

Since \( s \) is assumed to be stable and environment-invariant, we can decompose the prior \( p(c, t, s | Y, G) \) as:
\begin{equation}
p(c, t, s | Y, G) = p(c, t | Y, G) \, p(s | Y).
\end{equation}

With the above decomposition, the KL divergence term in Eq. \ref{eqn1} can be split into two parts. Consequently, we can obtain the final form of the ELBO as:
\begin{equation}
\begin{aligned}
& \mathcal{L}(X, Y, G)= \\
& \mathbb{E}_{q(c, t, s \mid X, Y, G)}\left[\log p(X \mid c, t, s, Y, G)\right]- \\
& \sum_i w_i\left(D_{\operatorname{KL}}\left(q\left(c, t \mid X_i, Y, G\right) \| p(c, t \mid Y, G)\right)+\right. \\
& D_{\mathrm{KL}}\left(q\left(s \mid X_i, Y, G\right) \| p(s \mid Y)\right) .
\end{aligned}
\end{equation}

To ensure that \( G \) denotes a valid DAG structure, we add an acyclicity constraint term \( \Upsilon(G) \), weighted by a coefficient \( \lambda \). Thus, the ELBO-transformed causal discovery optimization problem can be formulated as follows:
\begin{equation}
\min_{G} S(X, Y, G) = \mathcal{L}(X, Y, G) + \lambda \Upsilon(G) \quad \text{ s.t. } \quad G \in \text{DAG}.
\label{eqn2}
\end{equation}

Building on the above description, the transformed ELBO optimization formula integrates causal identifiability through variational inference, while ensuring DAG structure stability and robustness with an adaptive reweighting mechanism and acyclicity penalty. Furthermore, Eq. \ref{eqn2} indeed incorporates both inner and outer goals:
\vspace{0.75em}

\noindent \textbf{Inner Goal.} \emph{Enhance the model’s ability to capture key causal relationships by dynamically adjusting sample weights \( w_i \), focusing on the most informative samples to improve accuracy and robustness.}
\\

\noindent \textbf{Outer Goal.} \emph{Optimize the score function for the DAG structure \( G \), minimizing discrepancies between the learned and true causal relationships while preserving acyclicity.}
\vspace{0.75em}


To ensure HSCM converges to the true DAG, we introduce Theorem 2, which shows that finite sample weights do not affect the model's consistency or identifiability. Even with reweighted samples, the optimization process remains consistent, ensuring the learned DAG asymptotically converges to the true causal structure and preserves causal integrity over time. For the linear SEM in Theorem 2, the relationship can be written as \( X = X B + N \), where \( B \) represents the coefficient matrix and \( N \) denotes the noise term. For each observed sample \( x_i \) and each variable \( j \), we model:
\begin{equation}
X_{ij} = (X B){ij} + N{ij},
\end{equation}
where \( N_{ij} \sim \mathcal{N}(0, \sigma_j^2) \). Given Gaussian noise, the NLL for a single sample \( i \) and feature \( j \) is:
\begin{equation}
l_{ij} = -\log \left( \frac{1}{\sigma_j \sqrt{2\pi}} \right) + \frac{(X_{ij} - (X B){ij})^2}{2 \sigma_j^2}.
\end{equation}

The total weighted NLL loss is:
\begin{equation}
\begin{aligned}
& L_w(X, B)= \\
& \sum_{j=1}^d \sum_{i=1}^n w_i\left(-\log \left(\frac{1}{\sigma_j \sqrt{2 \pi}}\right)+\frac{\left(X_{i j}-(X B)_{i j}\right)^2}{2 \sigma_j^2}\right),
\end{aligned}
\end{equation}
where \( w_i \) are the sample weights that prioritize different observations, \( x_j \) is the \( j_{th} \) column of \( X \), \( \beta_j \) is the \( j_{th} \) column of \( B \), and \( W = \text{diag}(w_1, \dots, w_n) \) is the diagonal matrix of weights. To analyze the asymptotic convergence of the learned DAG to the ground-truth DAG, we use the negative log-likelihood loss with standard Gaussian noise \cite{peters2014causal, zheng2020learning, ng2022convergence}. To simplify, we focus only on the terms involving \( B \), as the factor \( \log \left( \frac{1}{\sigma_j \sqrt{2\pi}} \right) \) does not impact the optimization.
\begin{equation}
L_w(X, B) = \sum_{j=1}^d \sum_{i=1}^n \frac{w_i}{2 \sigma_j^2} (X_{ij} - (X B)_{ij})^2.
\end{equation}

Writing the loss in matrix form, we can get:
\begin{equation}
L_w(X, B) = \sum_{j=1}^d \frac{1}{2 \sigma_j^2} (x_j - X \beta_j)^\top W (x_j - X \beta_j).
\end{equation}

Expanding each term, the above equation can be formulated as follows:
\begin{equation}
\begin{aligned}
L_w(X, B) = & \sum_{j=1}^d \frac{1}{2 \sigma_j^2} ( x_j^\top W x_j - \\
& 2 x_j^\top W X \beta_j + \beta_j^\top X^\top W X \beta_j ).
\end{aligned}
\end{equation}

To find the optimal \( \beta_j \), we take the partial derivative of \( L_w(X, B) \) with respect to \( \beta_j \) and set it equal to zero:
\begin{equation}
\frac{\partial L_w(X, B)}{\partial \beta_j} = \frac{1}{2 \sigma_j^2} \left( -2 X^\top W x_j + 2 X^\top W X \beta_j \right) = 0.
\end{equation}

Simplifying, we can get:
\begin{equation}
X^\top W X \beta_j = X^\top W x_j.
\end{equation}

Thus, the solution for \( \beta_j \) is:
\begin{equation}
\beta_j = (X^\top W X)^{-1} X^\top W x_j.
\end{equation}

As \( n \to \infty \), the Law of Large Numbers implies that \( X^\top W X \) will converge to its expected value under the population distribution. Assuming \( W \) is chosen to reflect the population well, we obtain:
\begin{equation}
\lim_{n \to \infty} X^\top W X = \mathbb{E}[X^\top X].
\end{equation}

Similarly, \( X^\top W x_j \) converges to \( \mathbb{E}[X^\top x_j] \), so we have:
\begin{equation}
\lim_{n \to \infty} X^\top W x_j = \mathbb{E}[X^\top x_j].
\end{equation}

Thus, for sufficiently large \( n \), the weighted estimator \( \beta_j^{H} = (X^\top W X)^{-1} X^\top W x_j \) will approximate the standard estimator \( \beta_j = (X^\top X)^{-1} X^\top x_j \). This provides an independent proof that, asymptotically, the HSCM yields the same results as the standard approach when the sample size grows large, assuming appropriate weighting conditions.

\end{appendices}


\clearpage

\bibliography{ref}


\begin{thebibliography}{52}
\ifx \bisbn   \undefined \def \bisbn  #1{ISBN #1}\fi
\ifx \binits  \undefined \def \binits#1{#1}\fi
\ifx \bauthor  \undefined \def \bauthor#1{#1}\fi
\ifx \batitle  \undefined \def \batitle#1{#1}\fi
\ifx \bjtitle  \undefined \def \bjtitle#1{#1}\fi
\ifx \bvolume  \undefined \def \bvolume#1{\textbf{#1}}\fi
\ifx \byear  \undefined \def \byear#1{#1}\fi
\ifx \bissue  \undefined \def \bissue#1{#1}\fi
\ifx \bfpage  \undefined \def \bfpage#1{#1}\fi
\ifx \blpage  \undefined \def \blpage #1{#1}\fi
\ifx \burl  \undefined \def \burl#1{\textsf{#1}}\fi
\ifx \doiurl  \undefined \def \doiurl#1{\url{https://doi.org/#1}}\fi
\ifx \betal  \undefined \def \betal{\textit{et al.}}\fi
\ifx \binstitute  \undefined \def \binstitute#1{#1}\fi
\ifx \binstitutionaled  \undefined \def \binstitutionaled#1{#1}\fi
\ifx \bctitle  \undefined \def \bctitle#1{#1}\fi
\ifx \beditor  \undefined \def \beditor#1{#1}\fi
\ifx \bpublisher  \undefined \def \bpublisher#1{#1}\fi
\ifx \bbtitle  \undefined \def \bbtitle#1{#1}\fi
\ifx \bedition  \undefined \def \bedition#1{#1}\fi
\ifx \bseriesno  \undefined \def \bseriesno#1{#1}\fi
\ifx \blocation  \undefined \def \blocation#1{#1}\fi
\ifx \bsertitle  \undefined \def \bsertitle#1{#1}\fi
\ifx \bsnm \undefined \def \bsnm#1{#1}\fi
\ifx \bsuffix \undefined \def \bsuffix#1{#1}\fi
\ifx \bparticle \undefined \def \bparticle#1{#1}\fi
\ifx \barticle \undefined \def \barticle#1{#1}\fi
\bibcommenthead
\ifx \bconfdate \undefined \def \bconfdate #1{#1}\fi
\ifx \botherref \undefined \def \botherref #1{#1}\fi
\ifx \url \undefined \def \url#1{\textsf{#1}}\fi
\ifx \bchapter \undefined \def \bchapter#1{#1}\fi
\ifx \bbook \undefined \def \bbook#1{#1}\fi
\ifx \bcomment \undefined \def \bcomment#1{#1}\fi
\ifx \oauthor \undefined \def \oauthor#1{#1}\fi
\ifx \citeauthoryear \undefined \def \citeauthoryear#1{#1}\fi
\ifx \endbibitem  \undefined \def \endbibitem {}\fi
\ifx \bconflocation  \undefined \def \bconflocation#1{#1}\fi
\ifx \arxivurl  \undefined \def \arxivurl#1{\textsf{#1}}\fi
\csname PreBibitemsHook\endcsname

\bibitem[\protect\citeauthoryear{Yang et~al.}{2024}]{yang2024generalized}
\begin{barticle}
\bauthor{\bsnm{Yang}, \binits{J.}},
\bauthor{\bsnm{Zhou}, \binits{K.}},
\bauthor{\bsnm{Li}, \binits{Y.}},
\bauthor{\bsnm{Liu}, \binits{Z.}}:
\batitle{Generalized out-of-distribution detection: A survey}.
\bjtitle{Int. J. Comput. Vis.}
\bvolume{132}(\bissue{12}),
\bfpage{5635}--\blpage{5662}
(\byear{2024})
\end{barticle}
\endbibitem

\bibitem[\protect\citeauthoryear{Mart{\'\i}nez-S{\'a}nchez et~al.}{2024}]{martinez2024decomposing}
\begin{barticle}
\bauthor{\bsnm{Mart{\'\i}nez-S{\'a}nchez}, \binits{{\'A}.}},
\bauthor{\bsnm{Arranz}, \binits{G.}},
\bauthor{\bsnm{Lozano-Dur{\'a}n}, \binits{A.}}:
\batitle{Decomposing causality into its synergistic, unique, and redundant components}.
\bjtitle{Nat. Commun.}
\bvolume{15}(\bissue{1}),
\bfpage{9296}
(\byear{2024})
\end{barticle}
\endbibitem

\bibitem[\protect\citeauthoryear{Wang et~al.}{2022}]{wang2022generalizing}
\begin{barticle}
\bauthor{\bsnm{Wang}, \binits{J.}},
\bauthor{\bsnm{Lan}, \binits{C.}},
\bauthor{\bsnm{Liu}, \binits{C.}},
\bauthor{\bsnm{Ouyang}, \binits{Y.}},
\bauthor{\bsnm{Qin}, \binits{T.}},
\bauthor{\bsnm{Lu}, \binits{W.}},
\bauthor{\bsnm{Chen}, \binits{Y.}},
\bauthor{\bsnm{Zeng}, \binits{W.}},
\bauthor{\bsnm{Philip}, \binits{S.Y.}}:
\batitle{Generalizing to unseen domains: A survey on domain generalization}.
\bjtitle{IEEE Trans. Knowl. Data Eng.}
\bvolume{35}(\bissue{8}),
\bfpage{8052}--\blpage{8072}
(\byear{2022})
\end{barticle}
\endbibitem

\bibitem[\protect\citeauthoryear{Carlucci et~al.}{2019}]{carlucci2019domain}
\begin{bchapter}
\bauthor{\bsnm{Carlucci}, \binits{F.M.}},
\bauthor{\bsnm{D'Innocente}, \binits{A.}},
\bauthor{\bsnm{Bucci}, \binits{S.}},
\bauthor{\bsnm{Caputo}, \binits{B.}},
\bauthor{\bsnm{Tommasi}, \binits{T.}}:
\bctitle{Domain generalization by solving jigsaw puzzles}.
In: \bbtitle{Proc. IEEE Conf. Comput. Vis. Pattern Recognit. (CVPR)},
pp. \bfpage{2229}--\blpage{2238}
(\byear{2019})
\end{bchapter}
\endbibitem

\bibitem[\protect\citeauthoryear{Volpi et~al.}{2018}]{volpi2018generalizing}
\begin{bchapter}
\bauthor{\bsnm{Volpi}, \binits{R.}},
\bauthor{\bsnm{Namkoong}, \binits{H.}},
\bauthor{\bsnm{Sener}, \binits{O.}},
\bauthor{\bsnm{Duchi}, \binits{J.C.}},
\bauthor{\bsnm{Murino}, \binits{V.}},
\bauthor{\bsnm{Savarese}, \binits{S.}}:
\bctitle{Generalizing to unseen domains via adversarial data augmentation}.
In: \bbtitle{Adv. Neural Inf. Process. Syst. (NeurIPS)},
pp. \bfpage{1}--\blpage{11}
(\byear{2018})
\end{bchapter}
\endbibitem

\bibitem[\protect\citeauthoryear{Zhou et~al.}{2020a}]{zhou2020deep}
\begin{bchapter}
\bauthor{\bsnm{Zhou}, \binits{K.}},
\bauthor{\bsnm{Yang}, \binits{Y.}},
\bauthor{\bsnm{Hospedales}, \binits{T.}},
\bauthor{\bsnm{Xiang}, \binits{T.}}:
\bctitle{Deep domain-adversarial image generation for domain generalisation}.
In: \bbtitle{Proc. AAAI Conf. Artif. Intell. (AAAI)},
pp. \bfpage{13025}--\blpage{13032}
(\byear{2020})
\end{bchapter}
\endbibitem

\bibitem[\protect\citeauthoryear{Zhou et~al.}{2020b}]{zhou2020learning}
\begin{bchapter}
\bauthor{\bsnm{Zhou}, \binits{K.}},
\bauthor{\bsnm{Yang}, \binits{Y.}},
\bauthor{\bsnm{Hospedales}, \binits{T.}},
\bauthor{\bsnm{Xiang}, \binits{T.}}:
\bctitle{Learning to generate novel domains for domain generalization}.
In: \bbtitle{Proc. Europ. Conf. Comp. Vis. (ECCV)},
pp. \bfpage{561}--\blpage{578}
(\byear{2020})
\end{bchapter}
\endbibitem

\bibitem[\protect\citeauthoryear{Balaji et~al.}{2018}]{balaji2018metareg}
\begin{bchapter}
\bauthor{\bsnm{Balaji}, \binits{Y.}},
\bauthor{\bsnm{Sankaranarayanan}, \binits{S.}},
\bauthor{\bsnm{Chellappa}, \binits{R.}}:
\bctitle{Metareg: Towards domain generalization using meta-regularization}.
In: \bbtitle{Adv. Neural Inf. Process. Syst. (NeurIPS)},
pp. \bfpage{1}--\blpage{11}
(\byear{2018})
\end{bchapter}
\endbibitem

\bibitem[\protect\citeauthoryear{Li et~al.}{2019}]{li2019episodic}
\begin{bchapter}
\bauthor{\bsnm{Li}, \binits{D.}},
\bauthor{\bsnm{Zhang}, \binits{J.}},
\bauthor{\bsnm{Yang}, \binits{Y.}},
\bauthor{\bsnm{Liu}, \binits{C.}},
\bauthor{\bsnm{Song}, \binits{Y.-Z.}},
\bauthor{\bsnm{Hospedales}, \binits{T.M.}}:
\bctitle{Episodic training for domain generalization}.
In: \bbtitle{Proc. IEEE/CVF Int. Conf. Comput. Vis. (ICCV)},
pp. \bfpage{1446}--\blpage{1455}
(\byear{2019})
\end{bchapter}
\endbibitem

\bibitem[\protect\citeauthoryear{Chattopadhyay et~al.}{2020}]{chattopadhyay2020learning}
\begin{bchapter}
\bauthor{\bsnm{Chattopadhyay}, \binits{P.}},
\bauthor{\bsnm{Balaji}, \binits{Y.}},
\bauthor{\bsnm{Hoffman}, \binits{J.}}:
\bctitle{Learning to balance specificity and invariance for in and out of domain generalization}.
In: \bbtitle{Proc. Europ. Conf. Comp. Vis. (ECCV)},
pp. \bfpage{301}--\blpage{318}
(\byear{2020})
\end{bchapter}
\endbibitem

\bibitem[\protect\citeauthoryear{Piratla et~al.}{2020}]{piratla2020efficient}
\begin{bchapter}
\bauthor{\bsnm{Piratla}, \binits{V.}},
\bauthor{\bsnm{Netrapalli}, \binits{P.}},
\bauthor{\bsnm{Sarawagi}, \binits{S.}}:
\bctitle{Efficient domain generalization via common-specific low-rank decomposition}.
In: \bbtitle{Proc. Int. Conf. Mach. Learn. (ICML)},
pp. \bfpage{7728}--\blpage{7738}
(\byear{2020})
\end{bchapter}
\endbibitem

\bibitem[\protect\citeauthoryear{Dou et~al.}{2019}]{dou2019domain}
\begin{bchapter}
\bauthor{\bsnm{Dou}, \binits{Q.}},
\bauthor{\bsnm{Castro}, \binits{D.}},
\bauthor{\bsnm{Kamnitsas}, \binits{K.}},
\bauthor{\bsnm{Glocker}, \binits{B.}}:
\bctitle{Domain generalization via model-agnostic learning of semantic features}.
In: \bbtitle{Adv. Neural Inf. Process. Syst. (NeurIPS)}
(\byear{2019})
\end{bchapter}
\endbibitem

\bibitem[\protect\citeauthoryear{Wang et~al.}{2020}]{wang2020learning}
\begin{bchapter}
\bauthor{\bsnm{Wang}, \binits{S.}},
\bauthor{\bsnm{Yu}, \binits{L.}},
\bauthor{\bsnm{Li}, \binits{C.}},
\bauthor{\bsnm{Fu}, \binits{C.-W.}},
\bauthor{\bsnm{Heng}, \binits{P.-A.}}:
\bctitle{Learning from extrinsic and intrinsic supervisions for domain generalization}.
In: \bbtitle{Proc. Europ. Conf. Comp. Vis. (ECCV)},
pp. \bfpage{159}--\blpage{176}
(\byear{2020})
\end{bchapter}
\endbibitem

\bibitem[\protect\citeauthoryear{Chen et~al.}{2023}]{chen2023meta}
\begin{bchapter}
\bauthor{\bsnm{Chen}, \binits{J.}},
\bauthor{\bsnm{Gao}, \binits{Z.}},
\bauthor{\bsnm{Wu}, \binits{X.}},
\bauthor{\bsnm{Luo}, \binits{J.}}:
\bctitle{Meta-causal learning for single domain generalization}.
In: \bbtitle{Proc. IEEE Conf. Comput. Vis. Pattern Recognit. (CVPR)},
pp. \bfpage{7683}--\blpage{7692}
(\byear{2023})
\end{bchapter}
\endbibitem

\bibitem[\protect\citeauthoryear{Lv et~al.}{2022}]{lv2022causality}
\begin{bchapter}
\bauthor{\bsnm{Lv}, \binits{F.}},
\bauthor{\bsnm{Liang}, \binits{J.}},
\bauthor{\bsnm{Li}, \binits{S.}},
\bauthor{\bsnm{Zang}, \binits{B.}},
\bauthor{\bsnm{Liu}, \binits{C.H.}},
\bauthor{\bsnm{Wang}, \binits{Z.}},
\bauthor{\bsnm{Liu}, \binits{D.}}:
\bctitle{Causality inspired representation learning for domain generalization}.
In: \bbtitle{Proc. IEEE Conf. Comput. Vis. Pattern Recognit. (CVPR)},
pp. \bfpage{8046}--\blpage{8056}
(\byear{2022})
\end{bchapter}
\endbibitem

\bibitem[\protect\citeauthoryear{Ge et~al.}{2022}]{ge2022contributions}
\begin{bchapter}
\bauthor{\bsnm{Ge}, \binits{Y.}},
\bauthor{\bsnm{Xiao}, \binits{Y.}},
\bauthor{\bsnm{Xu}, \binits{Z.}},
\bauthor{\bsnm{Wang}, \binits{X.}},
\bauthor{\bsnm{Itti}, \binits{L.}}:
\bctitle{Contributions of shape, texture, and color in visual recognition}.
In: \bbtitle{Proc. Europ. Conf. Comp. Vis. (ECCV)},
pp. \bfpage{369}--\blpage{386}
(\byear{2022})
\end{bchapter}
\endbibitem

\bibitem[\protect\citeauthoryear{Wang et~al.}{2024}]{wang2024fusion}
\begin{botherref}
\oauthor{\bsnm{Wang}, \binits{Y.}},
\oauthor{\bsnm{Pedrycz}, \binits{W.}},
\oauthor{\bsnm{Ishibuchi}, \binits{H.}},
\oauthor{\bsnm{Zhu}, \binits{J.}}:
Fusion of explainable deep learning features using fuzzy integral in computer vision.
IEEE Trans. Fuzzy Syst.
(2024)
\end{botherref}
\endbibitem

\bibitem[\protect\citeauthoryear{Shi et~al.}{2022}]{shi2022visual}
\begin{botherref}
\oauthor{\bsnm{Shi}, \binits{B.}},
\oauthor{\bsnm{Song}, \binits{Y.}},
\oauthor{\bsnm{Joshi}, \binits{N.}},
\oauthor{\bsnm{Darrell}, \binits{T.}},
\oauthor{\bsnm{Wang}, \binits{X.}}:
Visual attention emerges from recurrent sparse reconstruction.
arXiv preprint arXiv:2204.10962,
1--19
(2022)
\end{botherref}
\endbibitem

\bibitem[\protect\citeauthoryear{Yang et~al.}{2025}]{yang2025dashboard}
\begin{botherref}
\oauthor{\bsnm{Yang}, \binits{M.}},
\oauthor{\bsnm{Hou}, \binits{Y.}},
\oauthor{\bsnm{Li}, \binits{L.}},
\oauthor{\bsnm{Chang}, \binits{R.}},
\oauthor{\bsnm{Zeng}, \binits{W.}}:
Dashboard vision: Using eye-tracking to understand and predict dashboard viewing behaviors.
IEEE Trans. Vis. Comput. Graph.,
1--16
(2025)
\end{botherref}
\endbibitem

\bibitem[\protect\citeauthoryear{Chen et~al.}{}]{chen2024frequency}
\begin{botherref}
\oauthor{\bsnm{Chen}, \binits{L.}},
\oauthor{\bsnm{Fu}, \binits{Y.}},
\oauthor{\bsnm{Gu}, \binits{L.}},
\oauthor{\bsnm{Yan}, \binits{C.}},
\oauthor{\bsnm{Harada}, \binits{T.}},
\oauthor{\bsnm{Huang}, \binits{G.}}:
Frequency-aware feature fusion for dense image prediction.
IEEE Trans. Pattern Anal. Mach. Intell.
\textbf{46},
10763--10780
\end{botherref}
\endbibitem

\bibitem[\protect\citeauthoryear{Wang et~al.}{2025}]{wang2025mmae}
\begin{barticle}
\bauthor{\bsnm{Wang}, \binits{X.}},
\bauthor{\bsnm{Fang}, \binits{L.}},
\bauthor{\bsnm{Zhao}, \binits{J.}},
\bauthor{\bsnm{Pan}, \binits{Z.}},
\bauthor{\bsnm{Li}, \binits{H.}},
\bauthor{\bsnm{Li}, \binits{Y.}}:
\batitle{Mmae: A universal image fusion method via mask attention mechanism}.
\bjtitle{Pattern Recognit.}
\bvolume{158},
\bfpage{111041}
(\byear{2025})
\end{barticle}
\endbibitem

\bibitem[\protect\citeauthoryear{Denison}{2024}]{denison2024visual}
\begin{barticle}
\bauthor{\bsnm{Denison}, \binits{R.N.}}:
\batitle{Visual temporal attention from perception to computation}.
\bjtitle{Nat. Rev. Psychol.}
\bvolume{3}(\bissue{4}),
\bfpage{261}--\blpage{274}
(\byear{2024})
\end{barticle}
\endbibitem

\bibitem[\protect\citeauthoryear{Zhou et~al.}{2022}]{zhou2022domain}
\begin{barticle}
\bauthor{\bsnm{Zhou}, \binits{K.}},
\bauthor{\bsnm{Liu}, \binits{Z.}},
\bauthor{\bsnm{Qiao}, \binits{Y.}},
\bauthor{\bsnm{Xiang}, \binits{T.}},
\bauthor{\bsnm{Loy}, \binits{C.C.}}:
\batitle{Domain generalization: A survey}.
\bjtitle{IEEE Trans. Pattern Anal. Mach. Intell.}
\bvolume{45}(\bissue{4}),
\bfpage{4396}--\blpage{4415}
(\byear{2022})
\end{barticle}
\endbibitem

\bibitem[\protect\citeauthoryear{Cubuk et~al.}{2020}]{cubuk2020randaugment}
\begin{bchapter}
\bauthor{\bsnm{Cubuk}, \binits{E.D.}},
\bauthor{\bsnm{Zoph}, \binits{B.}},
\bauthor{\bsnm{Shlens}, \binits{J.}},
\bauthor{\bsnm{Le}, \binits{Q.V.}}:
\bctitle{Randaugment: Practical automated data augmentation with a reduced search space}.
In: \bbtitle{Proc. IEEE Conf. Comput. Vis. Pattern Recognit. (CVPR)},
pp. \bfpage{702}--\blpage{703}
(\byear{2020})
\end{bchapter}
\endbibitem

\bibitem[\protect\citeauthoryear{Cubuk et~al.}{2019}]{cubuk2019autoaugment}
\begin{bchapter}
\bauthor{\bsnm{Cubuk}, \binits{E.D.}},
\bauthor{\bsnm{Zoph}, \binits{B.}},
\bauthor{\bsnm{Mane}, \binits{D.}},
\bauthor{\bsnm{Vasudevan}, \binits{V.}},
\bauthor{\bsnm{Le}, \binits{Q.V.}}:
\bctitle{Autoaugment: Learning augmentation strategies from data}.
In: \bbtitle{Proc. IEEE Conf. Comput. Vis. Pattern Recognit. (CVPR)},
pp. \bfpage{113}--\blpage{123}
(\byear{2019})
\end{bchapter}
\endbibitem

\bibitem[\protect\citeauthoryear{Xu et~al.}{2019}]{xu2019d}
\begin{bchapter}
\bauthor{\bsnm{Xu}, \binits{X.}},
\bauthor{\bsnm{Zhou}, \binits{X.}},
\bauthor{\bsnm{Venkatesan}, \binits{R.}},
\bauthor{\bsnm{Swaminathan}, \binits{G.}},
\bauthor{\bsnm{Majumder}, \binits{O.}}:
\bctitle{d-sne: Domain adaptation using stochastic neighborhood embedding}.
In: \bbtitle{Proc. IEEE Conf. Comput. Vis. Pattern Recognit. (CVPR)},
pp. \bfpage{2497}--\blpage{2506}
(\byear{2019})
\end{bchapter}
\endbibitem

\bibitem[\protect\citeauthoryear{Fan et~al.}{2021}]{fan2021adversarially}
\begin{bchapter}
\bauthor{\bsnm{Fan}, \binits{X.}},
\bauthor{\bsnm{Wang}, \binits{Q.}},
\bauthor{\bsnm{Ke}, \binits{J.}},
\bauthor{\bsnm{Yang}, \binits{F.}},
\bauthor{\bsnm{Gong}, \binits{B.}},
\bauthor{\bsnm{Zhou}, \binits{M.}}:
\bctitle{Adversarially adaptive normalization for single domain generalization}.
In: \bbtitle{Proc. IEEE Conf. Comput. Vis. Pattern Recognit. (CVPR)},
pp. \bfpage{8208}--\blpage{8217}
(\byear{2021})
\end{bchapter}
\endbibitem

\bibitem[\protect\citeauthoryear{Zhao et~al.}{2020}]{zhao2020maximum}
\begin{bchapter}
\bauthor{\bsnm{Zhao}, \binits{L.}},
\bauthor{\bsnm{Liu}, \binits{T.}},
\bauthor{\bsnm{Peng}, \binits{X.}},
\bauthor{\bsnm{Metaxas}, \binits{D.}}:
\bctitle{Maximum-entropy adversarial data augmentation for improved generalization and robustness}.
In: \bbtitle{Adv. Neural Inf. Process. Syst. (NeurIPS)},
pp. \bfpage{14435}--\blpage{14447}
(\byear{2020})
\end{bchapter}
\endbibitem

\bibitem[\protect\citeauthoryear{Koltchinskii}{2011}]{koltchinskii2011oracle}
\begin{bbook}
\bauthor{\bsnm{Koltchinskii}, \binits{V.}}:
\bbtitle{Oracle Inequalities in Empirical Risk Minimization and Sparse Recovery Problems: {\'E}cole D’{\'E}t{\'e} de Probabilit{\'e}s de Saint-Flour XXXVIII-2008}
vol. \bseriesno{2033},
(\byear{2011})
\end{bbook}
\endbibitem

\bibitem[\protect\citeauthoryear{Motiian et~al.}{2017}]{motiian2017unified}
\begin{bchapter}
\bauthor{\bsnm{Motiian}, \binits{S.}},
\bauthor{\bsnm{Piccirilli}, \binits{M.}},
\bauthor{\bsnm{Adjeroh}, \binits{D.A.}},
\bauthor{\bsnm{Doretto}, \binits{G.}}:
\bctitle{Unified deep supervised domain adaptation and generalization}.
In: \bbtitle{Proc. IEEE Int. Conf. Comput. Vis. (ICCV)},
pp. \bfpage{5715}--\blpage{5725}
(\byear{2017})
\end{bchapter}
\endbibitem

\bibitem[\protect\citeauthoryear{Wang et~al.}{2021}]{wang2021learning}
\begin{bchapter}
\bauthor{\bsnm{Wang}, \binits{Z.}},
\bauthor{\bsnm{Luo}, \binits{Y.}},
\bauthor{\bsnm{Qiu}, \binits{R.}},
\bauthor{\bsnm{Huang}, \binits{Z.}},
\bauthor{\bsnm{Baktashmotlagh}, \binits{M.}}:
\bctitle{Learning to diversify for single domain generalization}.
In: \bbtitle{Proc. IEEE/CVF Int. Conf. Comput. Vis. (ICCV)},
pp. \bfpage{834}--\blpage{843}
(\byear{2021})
\end{bchapter}
\endbibitem

\bibitem[\protect\citeauthoryear{Volpi and Murino}{2019}]{volpi2019addressing}
\begin{bchapter}
\bauthor{\bsnm{Volpi}, \binits{R.}},
\bauthor{\bsnm{Murino}, \binits{V.}}:
\bctitle{Addressing model vulnerability to distributional shifts over image transformation sets}.
In: \bbtitle{Proc. IEEE/CVF Int. Conf. Comput. Vis. (ICCV)},
pp. \bfpage{7980}--\blpage{7989}
(\byear{2019})
\end{bchapter}
\endbibitem

\bibitem[\protect\citeauthoryear{Jain et~al.}{2024}]{jain2024improving}
\begin{bchapter}
\bauthor{\bsnm{Jain}, \binits{N.}},
\bauthor{\bsnm{Suggala}, \binits{A.S.}},
\bauthor{\bsnm{Shenoy}, \binits{P.}}:
\bctitle{Improving generalization via meta-learning on hard samples}.
In: \bbtitle{Proc. IEEE Conf. Comput. Vis. Pattern Recognit. (CVPR)},
pp. \bfpage{27600}--\blpage{27609}
(\byear{2024})
\end{bchapter}
\endbibitem

\bibitem[\protect\citeauthoryear{Mahajan et~al.}{2021}]{mahajan2021domain}
\begin{bchapter}
\bauthor{\bsnm{Mahajan}, \binits{D.}},
\bauthor{\bsnm{Tople}, \binits{S.}},
\bauthor{\bsnm{Sharma}, \binits{A.}}:
\bctitle{Domain generalization using causal matching}.
In: \bbtitle{Proc. Int. Conf. Mach. Learn. (ICML)},
pp. \bfpage{7313}--\blpage{7324}
(\byear{2021})
\end{bchapter}
\endbibitem

\bibitem[\protect\citeauthoryear{Lin et~al.}{2024}]{lin2024towards}
\begin{bchapter}
\bauthor{\bsnm{Lin}, \binits{Y.}},
\bauthor{\bsnm{Zhao}, \binits{C.}},
\bauthor{\bsnm{Shao}, \binits{M.}},
\bauthor{\bsnm{Meng}, \binits{B.}},
\bauthor{\bsnm{Zhao}, \binits{X.}},
\bauthor{\bsnm{Chen}, \binits{H.}}:
\bctitle{Towards counterfactual fairness-aware domain generalization in changing environments}.
In: \bbtitle{Proc. Int. Jt. Conf. Artif. Intell. (IJCAI)},
pp. \bfpage{4560}--\blpage{4568}
(\byear{2024})
\end{bchapter}
\endbibitem

\bibitem[\protect\citeauthoryear{Jiao et~al.}{2025}]{jiao2025domain}
\begin{botherref}
\oauthor{\bsnm{Jiao}, \binits{P.}},
\oauthor{\bsnm{Zhao}, \binits{N.}},
\oauthor{\bsnm{Chen}, \binits{J.}},
\oauthor{\bsnm{Jiang}, \binits{Y.-G.}}:
Domain expansion and boundary growth for open-set single-source domain generalization.
IEEE Trans. Multimed.,
1--14
(2025)
\end{botherref}
\endbibitem

\bibitem[\protect\citeauthoryear{Li et~al.}{2021}]{li2021progressive}
\begin{bchapter}
\bauthor{\bsnm{Li}, \binits{L.}},
\bauthor{\bsnm{Gao}, \binits{K.}},
\bauthor{\bsnm{Cao}, \binits{J.}},
\bauthor{\bsnm{Huang}, \binits{Z.}},
\bauthor{\bsnm{Weng}, \binits{Y.}},
\bauthor{\bsnm{Mi}, \binits{X.}},
\bauthor{\bsnm{Yu}, \binits{Z.}},
\bauthor{\bsnm{Li}, \binits{X.}},
\bauthor{\bsnm{Xia}, \binits{B.}}:
\bctitle{Progressive domain expansion network for single domain generalization}.
In: \bbtitle{Proc. IEEE Conf. Comput. Vis. Pattern Recognit. (CVPR)},
pp. \bfpage{224}--\blpage{233}
(\byear{2021})
\end{bchapter}
\endbibitem

\bibitem[\protect\citeauthoryear{Mo et~al.}{2025}]{mo2025domain}
\begin{botherref}
\oauthor{\bsnm{Mo}, \binits{Z.}},
\oauthor{\bsnm{Zhang}, \binits{Z.}},
\oauthor{\bsnm{Tsui}, \binits{K.-L.}}:
Domain generalization study of empirical risk minimization from causal perspectives.
IEEE Trans. Multimed.,
1--13
(2025)
\end{botherref}
\endbibitem

\bibitem[\protect\citeauthoryear{Chattopadhay et~al.}{2018}]{chattopadhay2018grad}
\begin{bchapter}
\bauthor{\bsnm{Chattopadhay}, \binits{A.}},
\bauthor{\bsnm{Sarkar}, \binits{A.}},
\bauthor{\bsnm{Howlader}, \binits{P.}},
\bauthor{\bsnm{Balasubramanian}, \binits{V.N.}}:
\bctitle{Grad-cam++: Generalized gradient-based visual explanations for deep convolutional networks}.
In: \bbtitle{Proc. Int. Conf. Applications of Computer Vision (WACV)},
pp. \bfpage{839}--\blpage{847}
(\byear{2018})
\end{bchapter}
\endbibitem

\bibitem[\protect\citeauthoryear{LeCun et~al.}{1998}]{lecun1998gradient}
\begin{barticle}
\bauthor{\bsnm{LeCun}, \binits{Y.}},
\bauthor{\bsnm{Bottou}, \binits{L.}},
\bauthor{\bsnm{Bengio}, \binits{Y.}},
\bauthor{\bsnm{Haffner}, \binits{P.}}:
\batitle{Gradient-based learning applied to document recognition}.
\bjtitle{Proc. IEEE}
\bvolume{86}(\bissue{11}),
\bfpage{2278}--\blpage{2324}
(\byear{1998})
\end{barticle}
\endbibitem

\bibitem[\protect\citeauthoryear{Netzer et~al.}{2011}]{netzer2011reading}
\begin{bchapter}
\bauthor{\bsnm{Netzer}, \binits{Y.}},
\bauthor{\bsnm{Wang}, \binits{T.}},
\bauthor{\bsnm{Coates}, \binits{A.}},
\bauthor{\bsnm{Bissacco}, \binits{A.}},
\bauthor{\bsnm{Wu}, \binits{B.}},
\bauthor{\bsnm{Ng}, \binits{A.Y.}}:
\bctitle{Reading digits in natural images with unsupervised feature learning}.
In: \bbtitle{Adv. Neural Inf. Process. Syst. (NeurIPS)},
pp. \bfpage{1}--\blpage{8}
(\byear{2011})
\end{bchapter}
\endbibitem

\bibitem[\protect\citeauthoryear{Janzing and Sch{\"o}lkopf}{2010}]{janzing2010causal}
\begin{barticle}
\bauthor{\bsnm{Janzing}, \binits{D.}},
\bauthor{\bsnm{Sch{\"o}lkopf}, \binits{B.}}:
\batitle{Causal inference using the algorithmic markov condition}.
\bjtitle{IEEE Trans. Inf. Theory}
\bvolume{56}(\bissue{10}),
\bfpage{5168}--\blpage{5194}
(\byear{2010})
\end{barticle}
\endbibitem

\bibitem[\protect\citeauthoryear{Hull}{1994}]{hull1994database}
\begin{barticle}
\bauthor{\bsnm{Hull}, \binits{J.J.}}:
\batitle{A database for handwritten text recognition research}.
\bjtitle{IEEE Trans. Pattern Anal. Mach. Intell.}
\bvolume{16}(\bissue{5}),
\bfpage{550}--\blpage{554}
(\byear{1994})
\end{barticle}
\endbibitem

\bibitem[\protect\citeauthoryear{Krizhevsky and Hinton}{2009}]{krizhevsky2009learning}
\begin{botherref}
\oauthor{\bsnm{Krizhevsky}, \binits{A.}},
\oauthor{\bsnm{Hinton}, \binits{G.}}:
Learning multiple layers of features from tiny images,
1--60
(2009)
\end{botherref}
\endbibitem

\bibitem[\protect\citeauthoryear{Hendrycks and Dietterich}{2019}]{hendrycks2019benchmarking}
\begin{bchapter}
\bauthor{\bsnm{Hendrycks}, \binits{D.}},
\bauthor{\bsnm{Dietterich}, \binits{T.}}:
\bctitle{Benchmarking neural network robustness to common corruptions and perturbations}.
In: \bbtitle{Proc. Int. Conf. Learn. Represent. (ICLR)},
pp. \bfpage{1}--\blpage{16}
(\byear{2019})
\end{bchapter}
\endbibitem

\bibitem[\protect\citeauthoryear{Li et~al.}{2017}]{li2017deeper}
\begin{bchapter}
\bauthor{\bsnm{Li}, \binits{D.}},
\bauthor{\bsnm{Yang}, \binits{Y.}},
\bauthor{\bsnm{Song}, \binits{Y.-Z.}},
\bauthor{\bsnm{Hospedales}, \binits{T.M.}}:
\bctitle{Deeper, broader and artier domain generalization}.
In: \bbtitle{Proc. IEEE/CVF Int. Conf. Comput. Vis. (ICCV)},
pp. \bfpage{5542}--\blpage{5550}
(\byear{2017})
\end{bchapter}
\endbibitem

\bibitem[\protect\citeauthoryear{Venkateswara et~al.}{2017}]{venkateswara2017deep}
\begin{bchapter}
\bauthor{\bsnm{Venkateswara}, \binits{H.}},
\bauthor{\bsnm{Eusebio}, \binits{J.}},
\bauthor{\bsnm{Chakraborty}, \binits{S.}},
\bauthor{\bsnm{Panchanathan}, \binits{S.}}:
\bctitle{Deep hashing network for unsupervised domain adaptation}.
In: \bbtitle{Proc. IEEE Conf. Comput. Vis. Pattern Recognit. (CVPR)},
pp. \bfpage{5018}--\blpage{5027}
(\byear{2017})
\end{bchapter}
\endbibitem

\bibitem[\protect\citeauthoryear{Hinton and Roweis}{2002}]{hinton2002stochastic}
\begin{bchapter}
\bauthor{\bsnm{Hinton}, \binits{G.E.}},
\bauthor{\bsnm{Roweis}, \binits{S.}}:
\bctitle{Stochastic neighbor embedding}.
In: \bbtitle{Adv. Neural Inf. Process. Syst. (NeurIPS)},
pp. \bfpage{1}--\blpage{8}
(\byear{2002})
\end{bchapter}
\endbibitem

\bibitem[\protect\citeauthoryear{Rish et~al.}{2001}]{rish2001empirical}
\begin{bchapter}
\bauthor{\bsnm{Rish}, \binits{I.}}, \betal:
\bctitle{An empirical study of the naive bayes classifier}.
In: \bbtitle{Proc. Int. Jt. Conf. Artif. Intell. (IJCAI)},
pp. \bfpage{41}--\blpage{46}
(\byear{2001})
\end{bchapter}
\endbibitem

\bibitem[\protect\citeauthoryear{Peters et~al.}{2014}]{peters2014causal}
\begin{botherref}
\oauthor{\bsnm{Peters}, \binits{J.}},
\oauthor{\bsnm{Mooij}, \binits{J.M.}},
\oauthor{\bsnm{Janzing}, \binits{D.}},
\oauthor{\bsnm{Sch{\"o}lkopf}, \binits{B.}}:
Causal discovery with continuous additive noise models.
J. Mach. Learn. Res.,
1--45
(2014)
\end{botherref}
\endbibitem

\bibitem[\protect\citeauthoryear{Zheng et~al.}{2020}]{zheng2020learning}
\begin{bchapter}
\bauthor{\bsnm{Zheng}, \binits{X.}},
\bauthor{\bsnm{Dan}, \binits{C.}},
\bauthor{\bsnm{Aragam}, \binits{B.}},
\bauthor{\bsnm{Ravikumar}, \binits{P.}},
\bauthor{\bsnm{Xing}, \binits{E.}}:
\bctitle{Learning sparse nonparametric dags}.
In: \bbtitle{AISTATS},
pp. \bfpage{3414}--\blpage{3425}
(\byear{2020})
\end{bchapter}
\endbibitem

\bibitem[\protect\citeauthoryear{Ng et~al.}{2022}]{ng2022convergence}
\begin{bchapter}
\bauthor{\bsnm{Ng}, \binits{I.}},
\bauthor{\bsnm{Lachapelle}, \binits{S.}},
\bauthor{\bsnm{Ke}, \binits{N.R.}},
\bauthor{\bsnm{Lacoste-Julien}, \binits{S.}},
\bauthor{\bsnm{Zhang}, \binits{K.}}:
\bctitle{On the convergence of continuous constrained optimization for structure learning}.
In: \bbtitle{AISTATS},
pp. \bfpage{8176}--\blpage{8198}
(\byear{2022})
\end{bchapter}
\endbibitem

\end{thebibliography}

\end{document}